\documentclass[pdflatex,sn-mathphys-num]{sn-jnl}
\usepackage{geometry}
\usepackage{graphicx}%
\usepackage{multirow}%
\usepackage{amsmath,amssymb,amsfonts}%
\usepackage{amsthm}%
\usepackage[title]{appendix}%
\usepackage{cancel}%
\usepackage{textcomp}%
\usepackage{manyfoot}%
\usepackage{booktabs}%
\usepackage{algorithm}%
\usepackage{algorithmicx}%
\usepackage{algpseudocode}%
\usepackage{listings}%

\usepackage{comment}%
\usepackage[table]{xcolor}%

\theoremstyle{thmstyleone}%
\theoremstyle{thmstyletwo}%
\theoremstyle{thmstylethree}%
\definecolor{darkgreen}{HTML}{000000}
\definecolor{Yellow}{HTML}{F4CCCC}
\definecolor{LightSkyBlue}{HTML}{F4CCCC}
\definecolor{Peach}{HTML}{F4CCCC}

\begin{document}

\title[Article Title]{Cross-Modal Visuo-Tactile Object Perception}

\author*[1, 2]{\fnm{Anirvan} \sur{Dutta}}\email{anirvan.dutta95@gmail.com}
\author[3, 4]{\fnm{Simone} \sur{Tasciotti}}
\author[5, 6]{\fnm{Claudia} \sur{Cusseddu}}
\author[1, 7]{\fnm{Ang} \sur{Li}}
\author[4]{\fnm{Panayiota}
\sur{Poirazi}}
\author[5, 6]{\fnm{Julijana} \sur{Gjorgjieva}}
\author[2]{\fnm{Etienne} \sur{Burdet}}
\author[8,9]{\fnm{Patrick} \spfx{van der} \sur{Smagt}}
\author*[1,10]{\fnm{Mohsen} \sur{Kaboli}}\email{mohsen.kaboli@bmwgroup.com}

\affil[1]{\orgname{BMW Group AG}, \orgaddress{\city{München}, \country{Germany}}}
\affil[2]{\orgdiv{Department of Bioengineering}, \orgname{Imperial College of Science, Technology and Medicine}, \orgaddress{\city{London}, \country{UK}}}
\affil[3]{\orgdiv{Department of Biology}, \orgname{University of Crete}, \orgaddress{\city{Heraklion}, \country{Greece}}} %
\affil[4]{\orgname{Institute of Molecular Biology and Biotechnology, Foundation for Research and Technology-Hellas}, \orgaddress{\city{Heraklion}, \country{Greece}}} 
\affil[5]{\orgdiv{School of Medicine and Health, Institute for Neuroscience}, \orgname{Technical University of Munich}, \orgaddress{\city{Munich}, \country{Germany}}}
\affil[6]{\orgdiv{School of Life Sciences}, \orgname{Technical University of Munich}, \orgaddress{\city{Freising}, \country{Germany}}}
\affil[7]{\orgdiv{School of Computation, Information and Technology}, \orgname{Technical University of Munich}, \country{Germany}}
\affil[8]{\orgdiv{Faculty of Informatics}, \orgname{Eötvös Loránd University}, \orgaddress{\city{Budapest}, \country{Hungary}}}
\affil[9]{\orgname{Foundation Robotics Labs},\orgaddress{\city{\,München}, \country{Germany}}}
\affil[10]{\orgdiv{Electrical Engineering}, \orgname{Eindhoven University of Technology}, \country{Netherlands}}

\abstract{Estimating physical properties is critical for safe and efficient autonomous robotic manipulation, particularly during contact-rich interactions. In such settings, vision and tactile sensing provide complementary information about object geometry, pose, inertia, stiffness, and contact dynamics, such as stick-slip behavior. However, these properties are only indirectly observable and cannot always be modeled precisely (e.g., deformation in non-rigid objects coupled with nonlinear contact friction), making the estimation problem inherently complex and requiring sustained exploitation of visuo-tactile sensory information during action. Existing visuo-tactile perception frameworks have primarily emphasized forceful sensor fusion or static cross-modal alignment, with limited consideration of how uncertainty and beliefs about object properties evolve over time. Inspired by human multi-sensory perception and active inference, we propose the Cross-Modal Latent Filter (CMLF) to learn a structured, causal latent state-space of physical object properties. CMLF supports bidirectional transfer of cross-modal priors between vision and touch and integrates sensory evidence through a Bayesian inference process that evolves over time. Real-world robotic experiments demonstrate that CMLF improves the efficiency and robustness of latent physical properties estimation under uncertainty compared to baseline approaches. Beyond performance gains, the model exhibits perceptual coupling phenomena analogous to those observed in humans, including susceptibility to cross-modal illusions and similar trajectories in learning cross-sensory associations. Together, these results constitutes a significant step toward generalizable, robust and physically consistent cross-modal integration for robotic multi-sensory perception.}


\maketitle

\section{Introduction}
\label{sec:intro}
Deploying robotic systems in unstructured environments that demand contact-rich interactions, such as precision assembly or tool manipulation \cite{cui2021toward, billard2019trends}, requires the perception of latent physical properties beyond basic object recognition. Extrinsic properties (such as shape) and intrinsic properties (such as inertia and stiffness), govern the dynamics of interaction \cite{hollerbach89, uttayopas2023object}, and are essential to achieve robust and efficient manipulation. Estimating these properties in novel objects necessitates a perception framework that can integrate visual and tactile cues \cite{dutta2025predictive} effectively. Furthermore, in complex interaction scenarios, prior knowledge of an object’s intrinsic properties inferred from extrinsic visual cues can guide efficient interaction, while under visual occlusion or contact-rich conditions, local tactile feedback and inferred intrinsic properties can provide critical information about extrinsic attributes such as shape and pose, thereby improving manipulability. However, despite substantial progress in visual and tactile based robotic perception in recent decades, their synergistic interplay, central to the exemplary biological perception \cite{murraycrossmodal2023}, remains underexplored in robotics \cite{lee2020making, murali2022deep, fangbidirectional2024, duttacm, hengvitacformer2025, tvttrans25, caroleo2025cross}.

To design a framework for cross-modal visuo-tactile perception for robotic manipulation, we draw on findings from human sensorimotor perception showing that multi-sensory cues, including vision, tactile, and proprioceptive (haptics) information, interact synergistically to enhance object recognition and spatial localization beyond what any single modality can achieve \cite{ernst2002, ernst2004merging}. This multi-sensory integration forms a cornerstone of cognition and dexterous manipulation \cite{ etienneinteraction, cm_nature}. Fusing information across modalities yields estimates that outperform even the most accurate individual modality. While, in cases of conflict, one modality can bias the other, as demonstrated by the rubber hand illusion \cite{samad2015perception}. 

This behavior aligns with the principle of Bayesian cross-modal integration, where sensory modality specific and cross-modal signals are combined according to their relative reliability, enhancing robustness and perceptual salience under uncertainty \cite{steindevelopment2014}. Furthermore, cross-modal integration is predictive and context-sensitive: humans anticipate multi-sensory inputs, for example using visual cues to form priors that guide haptic exploration before contact \cite{fiorinineural2021, zeljko2022implicit, lucia2024effect}. Developmental studies show that such correspondences are learned in particular during infancy \cite{gori2008young} to yield robust forward models \cite{ernst2002, cesanek2021motor, brunelitdoes2015}, and can even lead to enhanced multi-sensory coupling as in synesthesia \cite{baroncohen1997synaesthesia}, underscoring their foundational role in human perception. These learned associations encode statistical links between vision and touch, playing a fundamental role in how humans anticipate object properties and execute precise manipulation \cite{cesanek2021motor}. Collectively, these findings indicate that cross-modality is fundamentally an intricate inference problem, rather than a process of simple sensory combination.

\begin{figure}[!t]
    \centering
    \includegraphics[width=\textwidth]{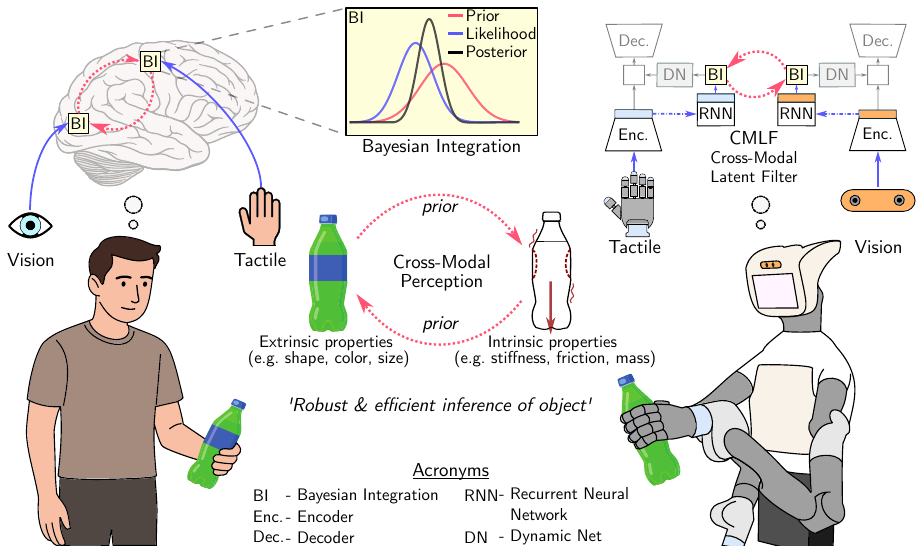}
    \caption{Concept of the proposed Cross-Modal Latent Filter (CMLF) perception framework. Inspired by human multi-sensory processing, the model uses extrinsic visual cues to form priors over intrinsic object properties, and vice versa, in a fully unsupervised manner from raw visual and tactile data. Bayesian integration (BI) underpins this cross-modal inference, improving the robustness and efficiency of object property estimation by exploiting statistical regularities across modalities.}
    \label{fig:graphicalabstract}
\end{figure}

How does the human brain integrate signals from individual sensory modalities into unified percepts and abstract representations that support inference itself, and generalize across novel situations \cite{pisella2001perception}? A key mechanism underlying this capability is \textit{active inference}: the notion that the brain generates percepts via top-down expectations guided by internal forward models, and refines them using sensory feedback through approximate Bayesian inference \cite{knillbayesian2004, friston2005theory, friston2009}. This active inference may be implemented at the cellular level~\cite{jordanconductancebased2024}, providing a mechanistic basis for uncertainty-aware computation. Motivated by these principles, we introduce a hierarchical deep state-space model \cite{latentmatters, dutta2025texture} that learns compact, low-dimensional latent representations from high-dimensional, spatio-temporal visuo-tactile observations, while explicitly incorporating top-down predictions within a variational inference scheme. A key contribution of our approach is the structured partitioning of the latent space into \textit{directly observable factors}, corresponding to sensory-driven features, and \textit{indirectly observable factors}, corresponding to object-level causal attributes. Within  this formulation, inference over intrinsic and extrinsic object properties is cast as a modality-specific Bayesian filtering problem, in which beliefs over latent factors are continuously updated as new sensory evidence is acquired during interaction. This yields a causal, object-centered latent state-space that captures both physical properties and their temporal evolution (dynamics) during manipulation. Explicit sequential inference overcomes the limitations of static methods under partial and temporally sparse observations, enabling uncertainty to decrease as evidence accumulates. This translates active inference from a neuroscience concept into a computationally grounded framework for robotic perception.

In addition, while human neurophysiological evidence indicates that cross-modal interactions can occur even within early sensory areas, the Bayesian integration and causal inference are primarily mediated by mid- to high-level cortical regions. This suggests that although primary cortices perform initial feature processing, these signals must converge in higher-order hubs to be weighted and integrated before feedback signals modulate activity across the hierarchy to shape a unified perception \cite{shamsearly2012, sebastianobalancing2024}. Converging evidence from mammalian models further supports this architectural organization~\cite{guyotoncortical2025, olcese2013multisensory}. In rodents, visuo-tactile integration is supported by well-characterized cortical circuits that encode abstract representations of peri-personal space—the near-body region in which vision and touch are naturally co-registered. In particular, associative posterior parietal areas, such as the rostro-lateral (RL) cortex, are anatomically positioned between primary visual (V1) and primary somatosensory (S1) cortices and contain neurons with spatially congruent visual and whisker receptive fields, enabling unified representations that support flexible behavior~\cite{wang2012network}. RL has been shown to be both necessary and sufficient for rapid cross-modal generalization of sensorimotor rules between touch and vision, consistent with the idea that higher-order cortex carries an abstract, modality-agnostic abstraction for near-body spatial variables \cite{guyotoncortical2025}.

At the microcircuit level, recordings in RL further indicate that neurons that appear unimodal at the spiking level can nonetheless receive subthreshold inputs from the other modality, implying latent multi-sensory structure that is not always expressed in overt output but can be recruited when reliability or context changes. This motivates preserving modality-specific inference pathways while enabling selective, alignment-based cross-modal coupling within a shared representational space~\cite{olcese2013multisensory}. Building on this principle, we introduce the \textit{cross-modal latent filter} (CMLF), which uses the structured latent partitioning (indirectly observable factors) to support flexible, bidirectional exchange of high-level information at the level of causal, object-centric properties, while maintaining modality-specific Bayesian inference. Cross-modal coupling is mediated through Bayesian integration, such that information from one modality constrains inference in the other exclusively via uncertainty-weighted priors over modality specific latent object properties, rather than rigid feature or latent alignments. This allows cross-modal associations to adapt dynamically as new interaction evidence becomes available, yielding a principled and robust framework for visuo-tactile object perception, mirroring biological evidence.

To systematically evaluate this cross-modal mapping, we design synthetic objects with extrinsic properties and intrinsic properties linked through explicit causal associations. Interacting with these objects potentially enables the learning of perceptual couplings that generalize classical human priors, such as the association between object size and weight \cite{cesanek2021motor}. We further introduce a test set of \textit{surprising objects} with deliberately inverted correspondences, allowing us to assess how the proposed CMLF learns and exploits such priors in an anticipatory manner, and to evaluate the robustness and adaptability of the resulting cross-modal associations.

Another principle in human neuroscience is that perception unfolds dynamically, with beliefs continuously updated over time through Bayesian integration of successive observations, a process critical for tactile perception, where contact information is sparse and intermittent \cite{roheneural2019}. However, much of the existing cross-modal robotics literature overlooks the temporal nature of perception, favoring static or single-step alignment schemes~\cite{murali2022deep, fangbidirectional2024, tvttrans25, caroleo2025cross}. These methods are typically evaluated on manually collected datasets and largely abstract away robotic interaction, focusing primarily on object recognition and thus offering limited applicability to downstream control tasks. Beyond temporal integration, recent studies also reveal that cortical circuits can extract shared abstract dynamics across modalities, allowing knowledge acquired through vision to transfer to touch, and vice versa \cite{guyotoncortical2025}. Inspired by these findings, our CMLF performs inference over time while learning shared interaction dynamics across modalities, overcoming the limitations of static methods under partial and temporally sparse observations, enabling uncertainty to decrease as evidence accumulates.

Developmental evidence provides additional insight into how such cross-modal structure emerges. A key prerequisite for multi-sensory integration is topographically co-aligned connectivity from multiple primary sensory cortices into associative targets. Recent work suggests a mechanism for how such co-alignment can arise early: correlated spontaneous activity in V1 and S1 before sensory experience can guide activity-dependent refinement of projections into RL, tuning the fraction of bimodal neurons and improving downstream decoding \cite{dwulet2024development}; moreover, these dynamics can be asymmetric, with somatosensory projections effectively ``leading'' visual projections during refinement. This offers a biological rationale for first stabilizing unimodal latent structure, then enabling cross-modal coupling once internally consistent representations exist \cite{murraycrossmodal2023}. Reflecting this principle, we train the model in an unsupervised manner, introducing cross-modal mapping only after initial unimodal learning, thereby mirroring the delayed and structured onset of cross-modal integration observed in biological systems.

The proposed framework, graphically represented in Figure~\ref{fig:graphicalabstract}, was systematically validated on a robotic platform comprising two manipulator arms equipped with vision and tactile sensors. CMLF brings biological-like cross-modality to robotic perception, advancing it along three key dimensions: (i) inference efficiency, (ii) robustness to noise and sensory corruption, and (iii) cross-modal perceptual coupling that mirrors biological perception, promoting improved generalization and offering insights into the developmental role of cross-modality in humans.
Conceptually, our design parallels a cortical motif in which V1- and S1-like streams construct modality-specific features, while a higher-level integration stage—analogous to association cortex—provides a shared representation that aligns touch and vision and supports cross-modal generalization during active sensing \cite{guyotoncortical2025, olcese2013multisensory}.

\section{Results}
\label{sec:results}
\begin{figure}[!tb]
    \centering
    \includegraphics[width=0.85\textwidth]{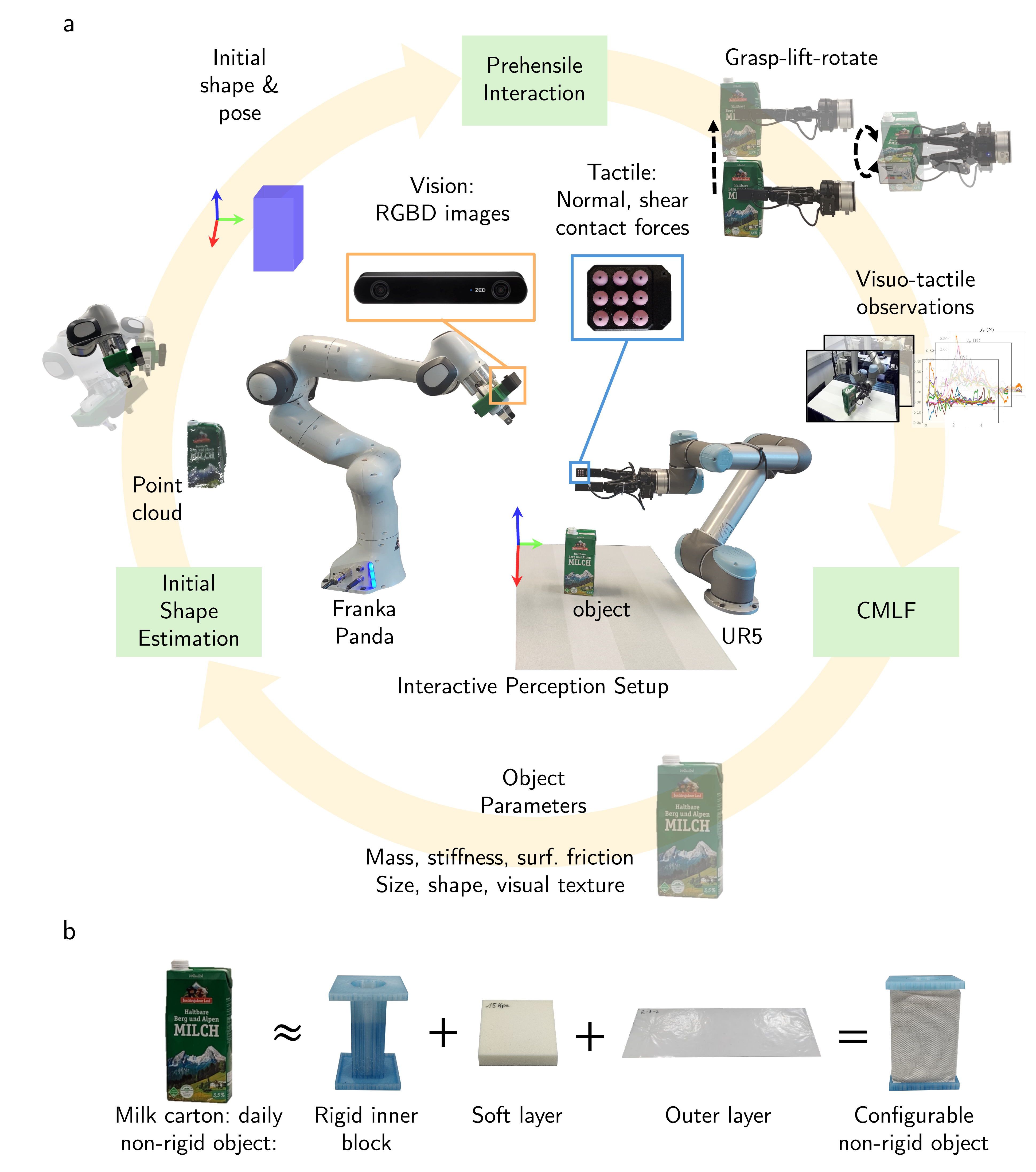}
     \caption{Robotic experimental setup and data collection pipeline. a) The figure illustrates the main stages used to construct the visuo-tactile interaction dataset. The process begins with initial estimation of the object shape and pose, which is used to autonomously perform prehensile robotic interactions. These interactions generate rich, time-varying visual and tactile observations capturing both geometric and contact dynamics. The resulting multi-sensory streams are then processed by the proposed Cross-Modal Latent Filter (CMLF) to infer latent object properties and interaction dynamics over time. b) To emulate everyday non-rigid objects, we designed a set of synthetic objects with configurable material properties, enabling controlled variation in flexibility. This allows us to systematically control intrinsic (e.g. mass, stiffness, surface friction) and extrinsic (e.g. size, shape, visual texture) physical attributes and to study their cross-modal associations.}
    \label{fig:experimentalsetuprobot}
\end{figure}
We first develop a mathematical framework for the proposed visuo-tactile cross-modality approach and then investigate how it supports the inference of physical properties of complex objects through experiments on our robotic platform (shown in Figure\,\ref{fig:experimentalsetuprobot}). Formally we address the problem of estimating the latent state $\mathbf{s}$ of an unknown, non-rigid object placed on a support surface from visual ($\mathbf{o}^V$) and tactile ($\mathbf{o}^T$) observations obtained through interactive actions $\mathbf{a}$. The interaction between objects and the robotic system is modeled as a discrete non-linear dynamical system with tactile and visual observations $\mathbf{o}_{1:H} = (\mathbf{o}_1, \mathbf{o}_2, \ldots, \mathbf{o}_H), \mathbf{o}_t \in \mathbb{R}^{n_o}$ and actions $\mathbf{a}_{1:H} = (\mathbf{a}_1, \mathbf{a}_2, \ldots, \mathbf{a}_H), \mathbf{a}_t \in \mathbb{R}^{n_a}$, defined over discrete time steps $t=1, \ldots, H$. At each time step, the observation vector is defined as the concatenation of visual and tactile measurements $\mathbf{o}_t = [\,\mathbf{o}^V_t,\ \mathbf{o}^T_t\,]'$ with $\mathbf{o}^V_t \in \mathbb{R}^{n^V_o}$ and $\mathbf{o}^T_t \in \mathbb{R}^{n^T_o}$, such that $n^{}_o = n^V_o + n^T_o$. 

The proposed approach begins with a shape exploration phase aimed at estimating the initial pose of the object and obtaining a compact, low-dimensional representation of its shape (see Methods~\ref{subsub:activeshape}). This is followed by a controlled prehensile interaction primitive (see Methods~\ref{subsub:prehensile}), designed to provide sufficient excitation of the object, yielding rich visuo-tactile observations. These observations are then processed by the proposed \textit{cross-modal latent filter} (CMLF), enabling the robotic system to infer the intrinsic and extrinsic properties of the object and estimate the latent state trajectory $\mathbf{s}_{1:H}$.

\subsection{Cross-Modal Latent Filter (CMLF)}
\label{subsec:cmlf}
We employ a \textit{deep state-space model} (DSSM)~\cite{karl2017deep} to capture interaction dynamics using low-dimensional latent variables $\mathbf{s}_{1:H} = (\mathbf{s}_{1}, \mathbf{s}_{2}, .., \mathbf{s}_{H})$, with $\mathbf{s}_{t} \in \mathbb{R}^{n_s}$ representing the system state at time $t$. The objective is to model the joint probability $p(\mathbf{o}_{1:H}, \mathbf{s}_{1:H}|\mathbf{a}_{1:H})$ and to perform variational inference by maximizing the likelihood of observations:
\begin{align}
    p(\mathbf{o}_{1:H}|\mathbf{a}_{1:H}) &= \! \int \!  p(\mathbf{o}_{1:H}, \mathbf{s}_{1:H}|\mathbf{a}_{1:H}) \, d\mathbf{s}_{1:H}.
\end{align}
To facilitate the analytical computation of the posterior while avoiding the computational expense associated with linearizing the observation model \cite{latentmatters}, we introduce a latent-space factorization motivated by the interaction structure and observability principles of observability of control theory~\cite{control1}. Specifically, the latent variable is decomposed as:
\begin{align}
    p(\mathbf{s}_t) = p(\mathbf{z}_t, \mathbf{y}_t) &= p(\mathbf{z}_t|\mathbf{y}_t) \, p(\mathbf{y}_t),
\end{align}
where $\mathbf{z}_t$ denotes directly observable components and $\mathbf{y}_t$ indirectly observable components, yielding: 
\begin{align}
   p(\mathbf{o}_{1:H}|\mathbf{a}_{1:H}) = \iint  \prod_{t=1}^H p(\mathbf{o}_t|\mathbf{z}_{t}) \, p(\mathbf{z}_t|\mathbf{z}_{t-1},\mathbf{y}_t, \mathbf{a}_t) \, p(\mathbf{y}_t|\mathbf{y}_{t-1}, \mathbf{a}_{t-1}) \, d\mathbf{y}_{t-1}  d\mathbf{z}_{t-1}.
\end{align}
In addition, the model employs separate latent spaces for the visual and tactile modalities, coupled through a shared dynamics model. This design enables the emergence of bidirectional cross-modal priors, allowing each modality to contribute complementary information during inference. Visual observations naturally encode extrinsic properties such as color and surface geometry, whereas tactile signals capture intricate intrinsic properties such as stiffness and compliance that are not visually observable.

\begin{figure}[!ht]
    \centering
    \includegraphics[width=0.9\textwidth]{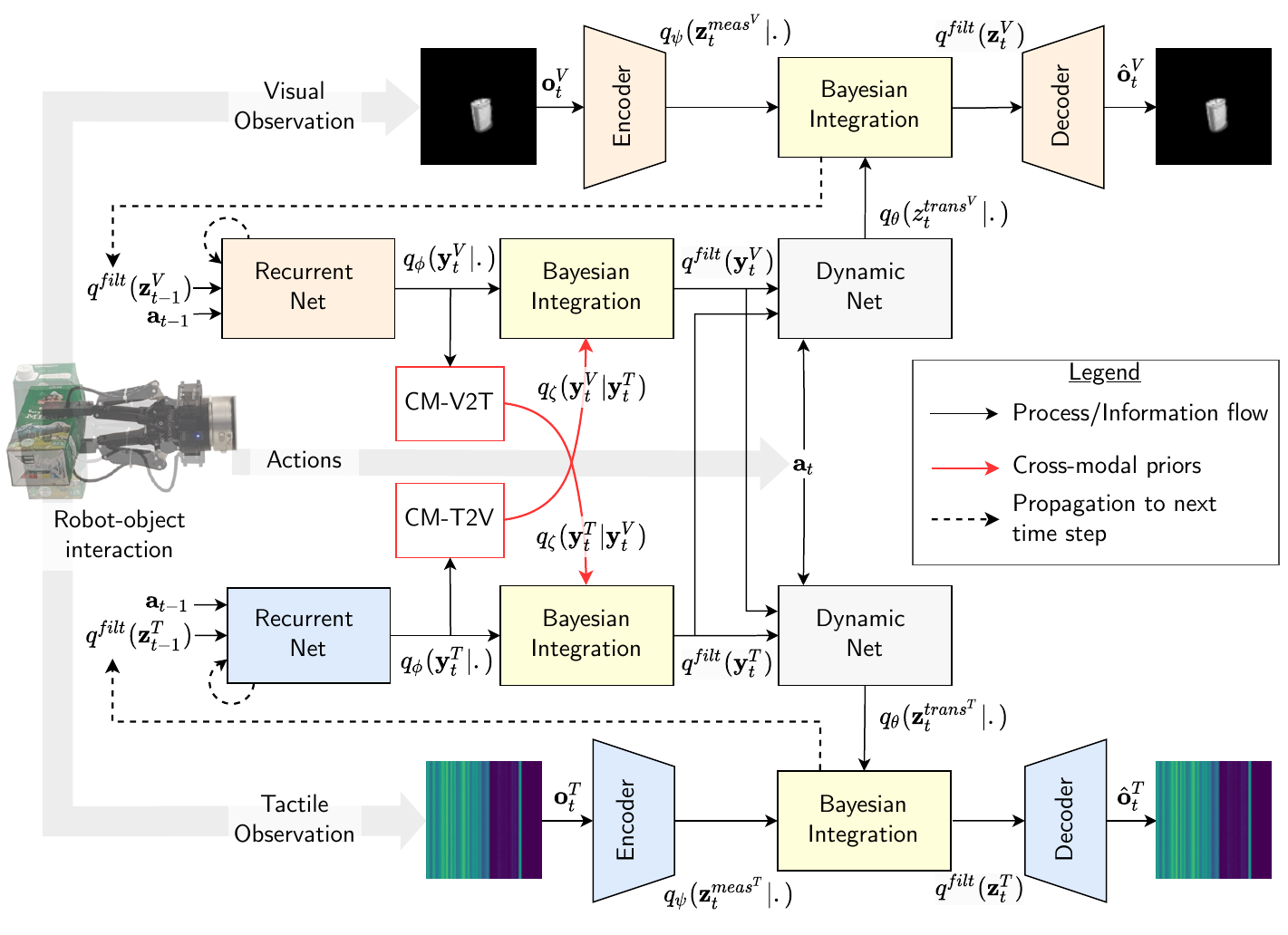}
    \caption{Cross-modal latent filter architecture with cross-modal connections from vision to tactile (\textsc{CM-V2T}) and tactile to vision (\textsc{CM-T2V}) modalities. The initial priors are set to $q^{filt}(\mathbf{z}_0^{V/T})\sim \mathcal{N}(0,1)$.}
    \label{fig:cmlatentframe}
\end{figure}

By structuring the latent representations into visual ($\mathbf{z}^{V}_t, \mathbf{y}^{V}_t$) and tactile ($\mathbf{z}^{T}_t, \mathbf{y}^{T}_t$) components, the model provides a principled mechanism for exchanging modality-specific priors during sequential inference. This improves robustness and efficiency when one modality is ambiguous, noisy, or unavailable (see the Supplementary, Section~1 for detailed derivation).
The \textit{evidence lower-bound objective} (ELBO) for the proposed CMLF is:
\begin{align}
    \mathcal{F}_{\textsc{ELBO}}(\theta, \phi, \psi, \zeta)  = \,\mathbb{E}_{q_\phi(.)}&\!\left[ \sum_{t=1}^{H} \log p(\mathbf{o}^{V}_t|\mathbf{z}^{V}_t)+\log p(\mathbf{o}^{T}_t|\mathbf{z}^{T}_t)\right] \nonumber \\
    - \sum_{t=2}^{H}{\textsc{KL}}\!\left[q^{filt}(\mathbf{z}^{V}_t)\,||\,p(\mathbf{z}^{V}_{t}|\mathbf{z}^{V}_{t-1}, \mathbf{y}^{V}_{t}, \mathbf{y}^{T}_{t}, \mathbf{a}_{t})\!\right] &- \sum_{t=2}^{H}{\textsc{KL}}\!\left[q^{filt}(\mathbf{z}^{T}_t)\,||\,p(\mathbf{z}^{T}_{t}|\mathbf{z}^{T}_{t-1}, \mathbf{y}^{V}_{t}, \mathbf{y}^{T}_{t}, \mathbf{a}_{t})\!\right] \nonumber \\
    -  \sum_{t=2}^{H}{\textsc{KL}}\!\left[q^{filt}(\mathbf{y}^{V}_t)\,||\,p(\mathbf{y}^{V}_t| \mathbf{a}_{t}, \mathcal{L})\!\right] &- \sum_{t=2}^{H}{\textsc{KL}}\!\left[q^{filt}(\mathbf{y}^{T}_t)\,||\,p(\mathbf{y}^{T}_t| \mathbf{a}_{t},\mathcal{L})\!\right].
    \label{eq:elbofinal}
\end{align}
Figure~\ref{fig:cmlatentframe} illustrates the overall architecture of the cross-model latent filter (CMLF). The inverse variational measurement models, $q_{\psi}(\mathbf{z}^{meas^{V/T}}_t|\mathbf{o}^{V/T}_t)$, are combined with the transition model $q_\theta(\mathbf{z}^{trans^{V/T}}_t|\mathbf{z}^{V/T}_{t-1}, \mathbf{y}^{V}_t, \mathbf{y}^{T}_t, \mathbf{a}_t)$ via Bayesian integration to compute the filtered variational distribution $q^{filt}(\mathbf{z}^{V/T}_t)$. The indirectly observable distribution $q_\phi(\mathbf{y}^{V/T}_t|\mathbf{z}^{V/T}_{t-1}, \mathbf{y}^{V/T}_{t-1}, \mathbf{a}_{t-1})$ is approximated using a \textit{long-short-term memory network} (LSTM) and combined with the cross-modal prior functions $q_\zeta(y^{V/T}_{t}|y^{T/V}_{t})$, through Bayesian integration to produce the filtered distribution $q^{filt}(\mathbf{y}^{V/T}_t)$.
For the directly observable space, the prior in each time step is taken as the prediction from the previous step. Additionally, a learnable hierarchical prior $p(\mathbf{y}^{T}_t|\mathbf{a}_{t}, \mathcal{L})$ is introduced to promote disentanglement and enforce contrastive behavior~\cite{flatvae}, where $\mathcal{L} \in \mathbb{N}$ denotes the object index serving as a physically grounded regularizer during CMLF training. Furthermore, cross-modal functions are activated only after the latent spaces of individual modalities have stabilized, the effect of which is studied and reported (see Results~\ref{subsec:humanalign}). The CMLF is trained by optimizing the ELBO in Eq.~\ref{eq:elbofinal}, with training and implementation details described in Section~\ref{subsec:preprocess}.


\subsection{CMLF improves inference efficiency}
\label{subsec:inference}
We evaluated the proposed Cross-Modal Latent Filter (CMLF) on a dataset of 75 systematically curated non-rigid objects (described in Section~\ref{subsec:expsetup}). To our knowledge there is no directly comparable state-of-the-art method that jointly performs bidirectional cross-modal inference and in dynamic settings. The approach most closely related \cite{fangbidirectional2024} is restricted to static scenarios and does not model the dependence of tactile observations on actions, making a direct quantitative comparison infeasible. To provide a meaningful point of reference, we therefore implemented a \textit{Baseline} approach based on a sequential variational autoencoder with a shared latent embedding \cite{chung2015recurrent}, a widely adopted architecture for multi-modal representation learning research.

In addition to this baseline, we conducted a systematic ablation study comprising (i) the proposed architecture and top-down feature extraction with hierarchical prior and joint space (\textit{Joint}), (ii) separate visual and tactile latent spaces \textit{without cross-modal connections} (wo-CM), and (iii) the full proposed approach with bidirectional \textit{cross-modal connections} (\textit{w-CM}). To analyze the learned latent representations, we trained a (five-fold cross-validated) logistic regression classifier on the latent features at the final time step (see Section~\ref{subsec:classreg}). 

\begin{figure}[!ht]
    \centering
    \includegraphics[width=\textwidth]{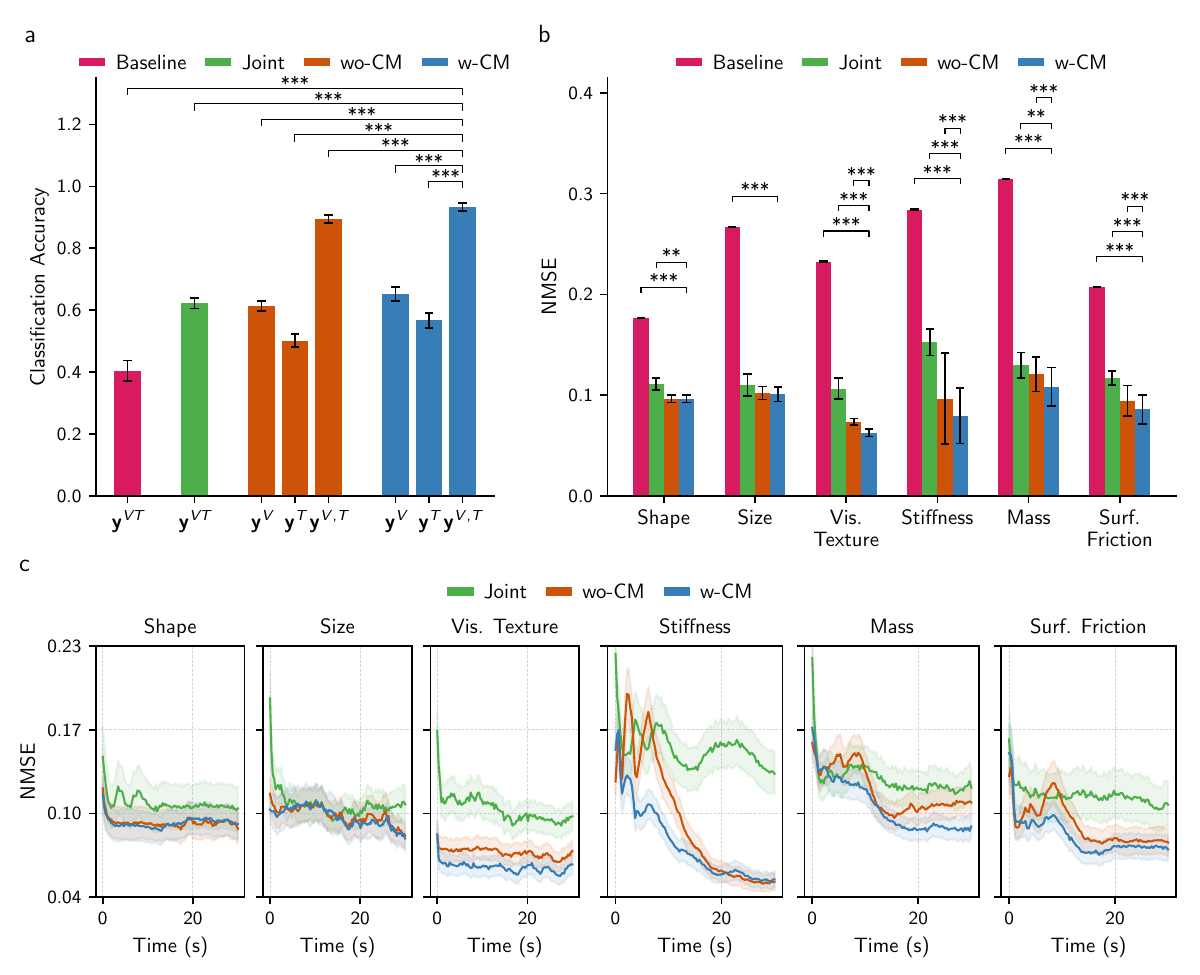}
    \caption{Classification and regression to evaluate inference performance. Statistical significance is assessed using paired t-tests with Holm--Bonferroni correction for multiple comparisons, with \textbf{***} denoting $p < 0.001$, \textbf{**} $p < 0.01$, \textbf{*} $p < 0.05$, and non-significant comparisons are omitted. (a) Classification accuracy for each latent feature with higher values indicate better performance and error bars denote $\pm\,1$ standard deviation. (b) Regression performance measured by trajectory NMSE for each physical property, with error bars denote $\pm\,1$ standard deviation; lower values indicate better estimation. (c) Temporal evolution of NMSE for intrinsic and extrinsic properties, with shaded regions indicating $\pm 0.1$ standard deviation across interactions; lower values indicate better estimation. The results show that cross-modal priors from extrinsic properties significantly improve the efficiency of intrinsic object property inference compared to baseline approaches.}
\label{fig:classregfig}
\end{figure}

We found that decomposing the latent space into separate visual and tactile components, introducing a hierarchy based on observability, and incorporating bidirectional cross-modal connections, yield highly discriminative features that can be separated by a linear decision boundary (Figure~\ref{fig:classregfig}\,a). Additionally, the visual modality ($y^{V}$) provides a slight advantage in object discrimination over the tactile modality ($y^{T}$) for this dataset, while combining both modalities improves classification performance. We also evaluated unsupervised metrics to qualitatively assess the learned latent structure, using pairwise Euclidean distances and 3D UMAP embeddings. However, due to the large number of objects, these visualizations were not sufficiently conclusive (Figure~2, see Supplementary, Section~2). Nevertheless, the relative trends and discriminative structure observed in the supervised classification analysis were consistently reflected in the unsupervised results.

A principled analysis was performed to determine whether the learned embeddings capture meaningful physical parameters rather than merely optimizing class separability. Since synthetically generated objects had intrinsic and extrinsic mechanical properties, the latent space representation was aligned with these ground-truth properties (see Section~\ref{subsec:classreg}). This alignment enabled analyzing the temporal evolution of physically grounded latent dimensions, providing deep insight into the interpretability of the representations and their effectiveness in inferring object properties. 

We found that the proposed CMLF (\textit{w-CM}) consistently outperformed all baseline approaches, achieving a lower mean inference error (Figure~\ref{fig:classregfig}b) and a higher inference efficiency for both intrinsic and extrinsic time-invariant properties. This efficiency is evidenced by faster convergence to the final estimate and reduced steady-state error over time (Figure~\ref{fig:classregfig}c). These results demonstrate that incorporating a cross-modal prior (\textit{w-CM}) substantially improves inferring efficiency for intrinsic properties such as stiffness, mass, and surface friction, as evidenced by a faster reduction in prediction error. In contrast, extrinsic properties show minimal early improvement, since visual observations dominate the initial 
exploratory phase, and tactile signals provide no useful prior at this stage.

Although the \textit{Baseline} approach can learn discriminative features, it lacks physical alignment, as reflected by the high regression errors (Figure~\ref{fig:classregfig}\,b), and was therefore excluded from further comparison. The \textit{Joint} variant performs noticeably better due to the separation of latent spaces and the hierarchical prior. However, its shared representation introduces ambiguity during later interaction phases—such as lifting and in-hand rotation—when both modalities contribute simultaneously. This leads to inconsistent latent structure and degraded predictions. In contrast, the proposed cross-modal approach (\textit{w-CM}) achieves the lowest estimation error (Figure~\ref{fig:classregfig}\,b) and the fastest convergence (Figure~\ref{fig:classregfig}\,c), highlighting the effectiveness of bilateral cross-modal connections within the CMLF.

Finally, we evaluated the ability of the model to infer pose trajectories varying in time (Figure~3; see Supplementary, Section~2). Despite the absence of explicit pose supervision, the resulting estimates are comparable to state-of-the-art supervised methods~\cite{hoque2021comprehensive}. This performance likely stems from the use of sequential modeling across both our approach and the baselines, combined with the relatively low level of occlusion or visual clutter in our experimental setup. Although pose estimation is not the primary focus of this work, these results highlight the robustness and broader applicability of the learned representations.


\subsection{CMLF is robust to noise and corruption}
\label{subsec:robustness}
Biologically, multi-sensory integration is most beneficial when evidence from individual modalities is weak or ambiguous - an effect classically described by \textit{inverse effectiveness}, and when information from one modality can compensate for degraded input in another. More generally, cortical circuits can carry cross-modal influences that are not always evident in spiking output under baseline conditions, but can become functionally important when sensory evidence is compromised—effectively providing a latent ``backup'' signal \cite{olcese2013multisensory}.

To this end, we conducted perturbation studies to evaluate whether cross-modal priors enhance inference robustness to observational noise and missing sensory data. Two types of perturbation were applied: (i) additive Gaussian white noise with variance $\sigma$, and (ii) observation corruption $c$, defined as the probability of dropout at any time step along the trajectory. Importantly, these perturbations were applied on the pre-processed input images and were not encountered during model training (see illustration of perturbation in Section~\ref{subsec:perturbation}.)

\begin{table}[!ht]
\centering
\caption{Perturbation study on noise and corruption, with heatmap in cells. Brighter hue signifies higher error and worst performance, values are scaled by 100 to improve clarity of presentation.}
\label{tab:noisecorr}
\begin{tabular*}{\textwidth}{@{\extracolsep\fill}ccccccc}
\toprule
\multicolumn{1}{l}{}           &          & \multicolumn{1}{c}{Nominal}        & \multicolumn{2}{c}{Noise}                                               & \multicolumn{2}{c}{Corruption}                                   \\
\cmidrule{3-7}
Parameters                     & Methods  & \multicolumn{1}{c}{$\sigma,c$=0} & \multicolumn{1}{c}{$\sigma$=0.2} & \multicolumn{1}{c}{$\sigma$=0.4} & \multicolumn{1}{c}{$c$=0.15} & \multicolumn{1}{c}{$c$=0.35}  \\
\midrule

\multirow{3}{*}{Shape} & Joint    & \cellcolor{Peach!43!white}11$\pm$1.1    & \cellcolor{Peach!42!white}11$\pm$1.3    & \cellcolor{Peach!45!white}11$\pm$1.6    & \cellcolor{Peach!75!white}12$\pm$0.86    & \cellcolor{Peach!99!white}13$\pm$0.82    \\
                               & wo-CM    & \cellcolor{Peach!3!white}9.7$\pm$0.34    & \cellcolor{Peach!49!white}11$\pm$0.78    & \cellcolor{Peach!56!white}12$\pm$0.78    & \cellcolor{Peach!42!white}11$\pm$0.49    & \cellcolor{Peach!50!white}11$\pm$0.45    \\
                               & w-CM    & \cellcolor{Peach!0!white}9.6$\pm$0.36     & \cellcolor{Peach!27!white}11$\pm$0.83     & \cellcolor{Peach!29!white}11$\pm$0.54     & \cellcolor{Peach!11!white}10$\pm$0.53     & \cellcolor{Peach!27!white}11$\pm$0.86     \\

\cmidrule{1-2}
\multirow{3}{*}{Size} & Joint    & \cellcolor{Peach!24!white}11$\pm$1.3    & \cellcolor{Peach!39!white}12$\pm$1.5    & \cellcolor{Peach!56!white}12$\pm$1.9    & \cellcolor{Peach!60!white}12$\pm$1.3    & \cellcolor{Peach!99!white}14$\pm$1.2    \\
                               & wo-CM    & \cellcolor{Peach!6!white}10$\pm$1.1    & \cellcolor{Peach!58!white}12$\pm$1.3    & \cellcolor{Peach!66!white}13$\pm$1.7    & \cellcolor{Peach!40!white}12$\pm$0.88    & \cellcolor{Peach!70!white}13$\pm$0.62    \\
                               & w-CM    & \cellcolor{Peach!0!white}10$\pm$1.5     & \cellcolor{Peach!27!white}11$\pm$0.94     & \cellcolor{Peach!46!white}12$\pm$0.91     & \cellcolor{Peach!21!white}11$\pm$1.3     & \cellcolor{Peach!51!white}12$\pm$1.3     \\

\cmidrule{1-2}
\multirow{3}{*}{Vis. Texture} & Joint    & \cellcolor{Peach!65!white}11$\pm$1.2    & \cellcolor{Peach!64!white}11$\pm$1.1    & \cellcolor{Peach!79!white}12$\pm$1.1    & \cellcolor{Peach!78!white}12$\pm$0.77    & \cellcolor{Peach!99!white}13$\pm$0.66    \\
                               & wo-CM    & \cellcolor{Peach!17!white}7.4$\pm$0.47    & \cellcolor{Peach!38!white}8.8$\pm$0.29    & \cellcolor{Peach!29!white}8.2$\pm$0.42    & \cellcolor{Peach!36!white}8.7$\pm$0.36    & \cellcolor{Peach!45!white}9.3$\pm$0.32    \\
                               & w-CM    & \cellcolor{Peach!0!white}6.3$\pm$0.49     & \cellcolor{Peach!19!white}7.5$\pm$0.39     & \cellcolor{Peach!24!white}7.9$\pm$0.58     & \cellcolor{Peach!16!white}7.3$\pm$0.42     & \cellcolor{Peach!21!white}7.7$\pm$0.5     \\

\cmidrule{1-2}
\multirow{3}{*}{Stiffness} & Joint    & \cellcolor{Peach!74!white}15$\pm$1.7    & \cellcolor{Peach!84!white}16$\pm$1.7    & \cellcolor{Peach!99!white}18$\pm$1.8    & \cellcolor{Peach!88!white}17$\pm$1.2    & \cellcolor{Peach!98!white}18$\pm$0.76    \\
                               & wo-CM    & \cellcolor{Peach!18!white}9.8$\pm$4.3    & \cellcolor{Peach!23!white}10$\pm$4.3    & \cellcolor{Peach!23!white}10$\pm$3.9    & \cellcolor{Peach!23!white}10$\pm$3.7    & \cellcolor{Peach!25!white}10$\pm$3.2    \\
                               & w-CM    & \cellcolor{Peach!0!white}7.9$\pm$2.8     & \cellcolor{Peach!4!white}8.4$\pm$3.4     & \cellcolor{Peach!13!white}9.3$\pm$3.6     & \cellcolor{Peach!3!white}8.3$\pm$2.7     & \cellcolor{Peach!15!white}9.5$\pm$2.6     \\

\cmidrule{1-2}
\multirow{3}{*}{Mass} & Joint    & \cellcolor{Peach!33!white}13$\pm$1.4    & \cellcolor{Peach!35!white}13$\pm$1.5    & \cellcolor{Peach!57!white}14$\pm$1.7    & \cellcolor{Peach!64!white}15$\pm$1.5    & \cellcolor{Peach!99!white}17$\pm$1.7    \\
                               & wo-CM    & \cellcolor{Peach!21!white}12$\pm$1.3    & \cellcolor{Peach!35!white}13$\pm$1.9    & \cellcolor{Peach!38!white}13$\pm$1.4    & \cellcolor{Peach!37!white}13$\pm$1.1    & \cellcolor{Peach!27!white}13$\pm$0.74    \\
                               & w-CM    & \cellcolor{Peach!0!white}11$\pm$1.6     & \cellcolor{Peach!2!white}11$\pm$1.7     & \cellcolor{Peach!9!white}11$\pm$1.2     & \cellcolor{Peach!7!white}11$\pm$1.4     & \cellcolor{Peach!17!white}12$\pm$1     \\

\cmidrule{1-2}
\multirow{3}{*}{Surf. Texture} & Joint    & \cellcolor{Peach!57!white}12$\pm$1.2    & \cellcolor{Peach!54!white}12$\pm$1.4    & \cellcolor{Peach!59!white}12$\pm$2    & \cellcolor{Peach!72!white}13$\pm$1.4    & \cellcolor{Peach!99!white}14$\pm$1.1    \\
                               & wo-CM    & \cellcolor{Peach!17!white}9.5$\pm$2.2    & \cellcolor{Peach!34!white}10$\pm$2.5    & \cellcolor{Peach!36!white}11$\pm$2.7    & \cellcolor{Peach!41!white}11$\pm$2.5    & \cellcolor{Peach!49!white}11$\pm$2.2    \\
                               & w-CM    & \cellcolor{Peach!0!white}8.6$\pm$2.3     & \cellcolor{Peach!4!white}8.8$\pm$2.7     & \cellcolor{Peach!15!white}9.4$\pm$3.1     & \cellcolor{Peach!1!white}8.7$\pm$1.8     & \cellcolor{Peach!12!white}9.3$\pm$1.6     \\

\cmidrule{1-2}
\multirow{3}{*}{Overall} & Joint    & \cellcolor{Peach!54!white}12$\pm$3    & \cellcolor{Peach!58!white}12$\pm$3    & \cellcolor{Peach!72!white}13$\pm$3    & \cellcolor{Peach!75!white}13$\pm$3.1    & \cellcolor{Peach!99!white}15$\pm$3.4    \\
                               & wo-CM    & \cellcolor{Peach!15!white}9.8$\pm$2.5    & \cellcolor{Peach!36!white}11$\pm$2.7    & \cellcolor{Peach!37!white}11$\pm$2.6    & \cellcolor{Peach!35!white}11$\pm$2.4    & \cellcolor{Peach!40!white}11$\pm$2.3    \\
                               & w-CM    & \cellcolor{Peach!0!white}8.9$\pm$2.4     & \cellcolor{Peach!11!white}9.6$\pm$2.6     & \cellcolor{Peach!20!white}10$\pm$2.4     & \cellcolor{Peach!9!white}9.4$\pm$2.3     & \cellcolor{Peach!22!white}10$\pm$2.4     \\

\end{tabular*}
\end{table}

The mean estimation error across interactions for different methods, evaluated across multiple physical properties and varying levels of noise and corruption demonstrate that the cross-modality integration significantly improves the model’s robustness to both noise and missing observations in the raw sensory inputs (Table~\ref{tab:noisecorr}). Notably, the prior connection from the tactile to the visual modality provides a clear benefit, which was less pronounced when focusing solely on inference efficiency (Sec.\ref{subsec:inference}). The proposed approach with cross-modal priors (\textit{w-CM}) consistently achieves lower prediction errors than the variant without cross-modal integration (\textit{wo-CM}) and significantly outperforms the \textit{Joint} approach.

\subsection{CMLF exhibits cross-modal coupling consistent with biological systems}
\label{subsec:humanalign}

From a biological point of view, cross-modal coupling in the cortex, especially in rodent systems, appears to be shaped by the maturation of unimodal representations and by the emergence of topographically aligned projections in the associative cortex \cite{dwulet2024development}. Developmental research suggests that correlated spontaneous activity can refine co-aligned projections from V1/S1 to area RL, prior to the onset of overt multi-sensory behavior \cite{dwulet2024development}, implying that a stable unimodal structure may be a prerequisite for effective cross-modal binding. This motivates testing whether delaying cross-modal coupling until the unimodal latent-space stabilizes improves learning. 

\begin{figure}[!th]
    \centering
    \includegraphics[width=\textwidth]{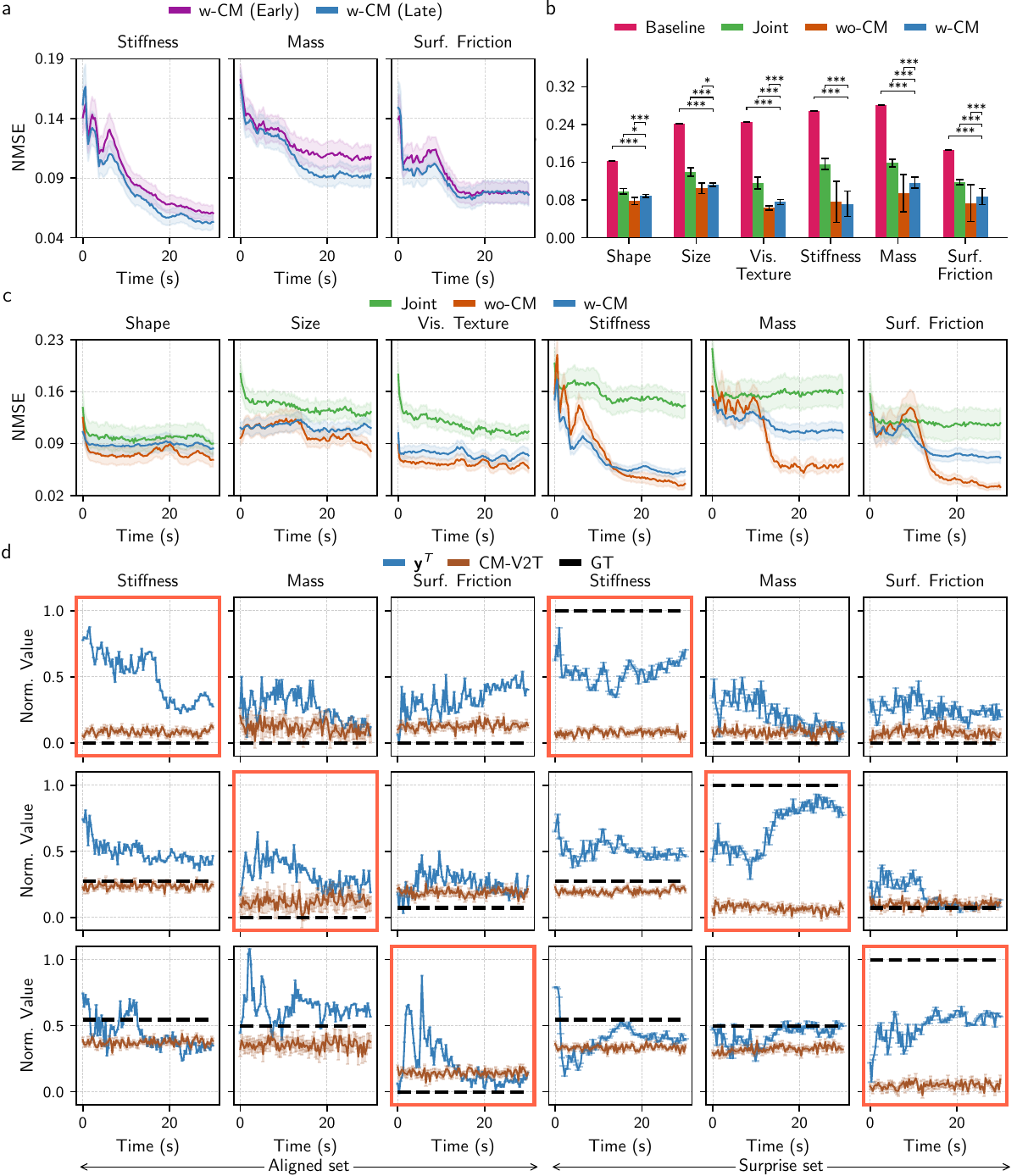}
    \caption{CMLF perceptual similarity to human inference. a) Mean trajectory error on the \textit{surprise set}. Statistical significance is evaluated using paired t-tests with Holm--Bonferroni correction. b) Effect of delayed activation of cross-modality priors on inference of intrinsic properties.  c) Temporal evolution of NMSE on the \textit{surprise set}, with shaded regions indicating $\pm 0.1$ standard deviation. d) Representative examples illustrating how latent filtering and cross-modal priors differ between the aligned and surprise sets. In the aligned set, the visual-to-tactile prior ($CM-V2T$) provides informative cues derived from extrinsic properties, enabling faster convergence of latent estimates. With the \textit{surprise set}, the prior trained on \textit{aligned set's} cross-modal association becomes misleading; however Bayesian integration gradually corrects the estimate toward the ground-truth.}
    \label{fig:cmbiologicalconn}
\end{figure}

Therefore, we performed a comparative analysis in which two conditions are considered: $w$-$CM(Late)$ denotes activation of the cross-modal priors after 25\% of the training epochs, while $w$-$CM(Early)$ refers to activation at 10\% of the training epochs. The results demonstrate that the timing of the cross-modal prior activation significantly affects the performance of the model, with more effective learning achieved when latent representations are allowed to stabilize before introducing cross-modal connections (Figure~\ref{fig:cmbiologicalconn},a). To our knowledge, this effect has not been explicitly examined in previous works, and this observation may also offer potential insight into biological cross-modal learning mechanisms, a topic that continues to be actively explored in neuroscience and cognitive science research.

Furthermore, biological multi-sensory systems respond asymmetrically to cross-modal structure, generalizing rapidly under consistent prior and likelihood (rule-preserving switches), but becomes slow when previously learned cross-modal correspondences are violated (rule-reversing switches), reflecting the need to override an internalized prior \cite{guyotoncortical2025}. This phenomenon is well illustrated in rodents by the peri-personal space circuit centered on area RL, where disruption of the circuit abolishes the ``pre-loaded'' cross-modal structure and forces relearning from unimodal inputs. Inspired by this, we constructed `surprising' objects that invert learned visuo-tactile correspondences to probe whether our learned cross-modal priors behave like biological priors, initially biasing inference, but being updated as sequential evidence is integrated. Six such objects were created to challenge the learned representations, including a stiffer cylinder, a softer hexagonal prism, a low-friction black object, a high-friction white object, and a short object with greater mass than a taller one (Figure~\ref{fig:experimentalsetupcm}\,c).

Comparing the mean error and the temporal evolution of the inference error in the surprise set, respectively, reveal a drop in inference performance, particularly for intrinsic properties, when using the cross-modal approach (\textit{w-CM}) compared to the \textit{wo-CM} (Figure~\ref{fig:cmbiologicalconn}\,b,c). The primary reason for this decline is that, in the aligned set, the visual-to-tactile prior ($CM-V2T$) provides useful cues linked to the intrinsic properties from the initial extrinsic observations, enabling the latent estimates to stabilize more rapidly to the true value. Under the surprise set, however, this same prior becomes unreliable because it reflects contradicting correlations learned only from the aligned set. To further analyze this effect, we examined instances in which surprising visual priors ($CM$-$V2T$) biased the convergence of intrinsic estimates derived from extrinsic observations. Three representative examples illustrate the temporal evolution of the latent states for both the aligned and surprise sets (Figure~\ref{fig:cmbiologicalconn}d), depicting surprises in: (i) shape–stiffness, (ii) size–mass, and (iii) visual texture–surface friction, respectively (highlighted axis). We found that in the surprise set, the prior (blue) can initially steer estimates (brown) away from the true values (dotted black) because it predicts stiffness, mass, or friction based on correlations with extrinsic features of shape, size and visual texture learned during training.

Nevertheless, our results also demonstrate that the model continues to rely on cross-modal structure even under distributional shift in observation, with the Bayesian integration playing an important role for gradually refining the estimates over time. This is consistent with perceptual behaviors observed in humans~\cite{ernst2002, cesanek2021motor}, where prior expectations derived from one modality can bias judgments in another (as in the size–weight illusion), but inference remains anchored by a principled combination of prior and sensory evidence via Bayesian integration. One key difference is that humans can rapidly revise or down-weight learned priors when they conflict with current sensory evidence, whereas deep neural networks generally struggle to adapt their weights online. Although alternative mechanisms for online adaptation exist, such as those explored in our previous work~\cite{duttacm}, these often introduce substantial computational overhead and exhibit limited generalizability.

\section{Discussion}
\label{sec:discuss}
In this paper, we introduced a novel cross-modal visuo-tactile perception framework to address the challenge of inferring non-rigid object properties in a fully unsupervised manner. At its core, the proposed Cross-Modal Latent Filter (CMLF), an action‑conditioned deep state‑space model that encodes interaction dynamics into a structured latent manifold for causal inference of object properties. Prior cross-modal visuo-tactile robotic approaches operated in static or quasi-static settings, limiting cross-modal transfer to fixed regularities observed in passive observations. Furthermore, majority of these approaches assumed unidirectional transfer between modalities (e.g., vision-to-tactile or tactile-to-vision), or neglected the action-conditioned nature of tactile sensing altogether. As a result, these methods provided limited support for inferring indirectly observable intrinsic properties, which only could be inferred through dynamical interaction, and going beyond object recognition.

A key contribution of the CMLF lies in partitioning the latent space into directly and indirectly observable factors, enabling more interpretable inference of physical properties. This structure establishes principled cross-modal coupling between vision and tactile modalities through Bayesian integration, allowing information from one modality to influence the other through a probabilistic prior. As a result, cross-modal signals are weighted by their relative uncertainty rather than enforced through rigid latent alignments, paves the path for the model to update or relearn statistical associations as new interaction evidence becomes available. Incorporating a hierarchical prior further improved performance, as evidenced by the superior \textit{Joint} model relative to its ablative \textit{Baseline} variant. Importantly, this prior operates only during training and does not restrict generative flexibility during inference, in contrast to rigid class-conditioning approaches used in discriminative models.

CMLF can be viewed as an algorithmic analogue of an associative-cortex motif, in which modality-specific streams converge onto an abstract latent manifold that supports cross-modal transfer. In rodents, a single dorsal associative area (RL) can be necessary and sufficient for cross-modal generalization based on peri-personal space, and bidirectional interactions between sensory and associative cortices appear critical for maintaining consistent structure across modalities \cite{guyotoncortical2025}. In this interpretation, our bidirectional latent priors in CMLF play a role analogous to biologically induced structure, enabling rapid transfer when cross-modal statistics are aligned.
\\

\noindent \textit{Cross-modal integration for improved efficiency and robustness}\\
To enable a systematic investigation of the efficacy of the CMLF, we curated a set of novel configurable non-rigid objects whose extrinsic and intrinsic physical properties could be varied independently. Real-robot experiments using these objects demonstrated that the proposed framework substantially improves inference of both intrinsic (e.g., stiffness, mass, compliance, friction) and extrinsic (e.g., shape, size, visual texture) properties compared to state of the art sequential unsupervised baselines. The open-source dataset from these experiments provides the foundation for future cross-modal studies that involve shape, stiffness, friction, mass distribution, and texture.

Importantly, the biologically inspired cross-modal mapping serves a dual purpose: (i) visual cues provide informative priors that improve the efficiency of intrinsic property inference, and (ii) tactile cues enhance the robustness of extrinsic property inference under noise and corruption. Notably, cross-modal coupling yields the largest performance gains under degraded inputs, consistent with classical multisensory principles in which congruent cross-modal cues preferentially enhance weak signals \cite{guyotoncortical2025, olcese2013multisensory} \cite{guyotoncortical2025, olcese2013multisensory}. At the circuit level, congruent and incongruent cues combinations can induce enhancement or suppression, potentially mediated by local inhibitory circuitry (e.g., PV inter-neurons) in associative cortex. This provides a mechanistic hypothesis for when cross-modal priors should be amplified versus gated. 
\\

\noindent \textit{Biologically plausible cross-modal coupling}\\
Another key biological inspiration of the CMLF is its top-down predictive modeling (starting with a prior), which  essentially would allow each modality to operate at different sampling rates, with flexible handling of delays or missing observations. This functionality is not present in existing multi-sensory integration approaches~\cite{guo2025multi} that assume synchronous fusion or strong architectural coupling. By leveraging Bayesian integration over sequential dynamics, the model retains modality-specific pathways while at the same time benefiting from cross-modal interactions when these are informative. Beyond inference performance, we explored the effect of cross-modal connections to novel `surprising' objects in inference. These experiments revealed parallels with neuroscience findings, including Bayesian-like weighting of cues. Unlike humans, who rapidly adjust priors when evidence shifts, deep networks struggle with such online adaptation of learned correlation/prior. This highlights an important open challenge: developing more flexible, biologically inspired models for adaptive cross-modal inference. 

Nevertheless, our findings suggest that the proposed framework not only advances robotic perception, but also provides insight into the computational principles underlying biological cross-modal inference. In particular, the biologically inspired cross-modal perceptual coupling improves generalization while unexpectedly revealing parallels with the developmental nature of cross-modality in biological systems, an emergent behaviour that becomes evident through training and evaluation of the model. Our observation that delayed activation of cross-modal priors improves learning resonates with a developmental scenario: multi-sensory integration requires topographic co-alignment of projections from different sensory areas into higher-order cortex, and such alignment may be scaffolded by correlated spontaneous activity before sensory experience \cite{dwulet2024development}. This suggests a normative reason to organize learning by first stabilizing unimodal representations, then forming cross-modal couplings once latent spaces become internally consistent.
\\

\noindent \textit{Limitations}\\
To our knowledge, this work presents one of the first demonstration of unsupervised inference of physical properties on a real robotic platform using visuo-tactile sensing. Beyond perception, the learned latent representations provide a strong foundation for downstream control, supporting both classical controllers (e.g., slip-aware feedback grounded in tactile latent states) and learning-based manipulation policies that benefit from anticipatory and fine-grained motor control. Crucially, our results highlight the value of cross-modal representations: learned associations between intrinsic properties (e.g., material or compliance) and extrinsic properties (e.g., shape and pose) act as informative priors that enable more efficient and robust interaction than unimodal perception or the absence of prior structure. Such priors can bias haptic exploration, support pre-adaptation of motor commands prior to contact, and improve the stability and accuracy of online inference during manipulation. In this way, cross-modal learning not only enhances perception but also more tightly couples sensing and action, moving robotic interaction closer to the anticipatory and adaptive behavior observed in human manipulation.

At the same time, the surprise-object experiments reveal an important limitation: when learned visuo-tactile correspondences are violated, cross-modal priors can become maladaptive and must be revised or overridden. In biological systems, such mismatches can be relearned, often with characteristic timescales and dependence on higher-order circuitry \cite{guyotoncortical2025}. Addressing this limitation points to the need for more flexible mechanisms to update and modulate priors, rather than relying on fixed cross-modal couplings.

Beyond cross-modal structure, the directly observable tactile latent space (e.g., the $\mathbf{z}$-space) offers opportunities to track geometric deformations and other transient physical events during interaction. In the present study, we downsampled visual inputs and reformulated inherently low-resolution tactile observations to reduce computational cost and training time. Although sufficient for demonstrating the proposed framework, this design choice trades off information fidelity. Future work could use higher-resolution sensing and more powerful computation to better preserve spatial detail, further improving the accuracy and robustness of cross-modal inference.

An inherent assumption in our current setup is that the underlying event or cause is aligned across modalities. Extending the framework toward causal inference, where cross-modal connections are activated only when modalities share a common cause, would move the model closer to Bayesian causal inference observed in human sensorimotor apparatus~\cite{heald2021contextual}. More generally, this framing captures an important qualitative feature of multi-sensory perception: congruent cues tend to be integrated and amplified, whereas incongruent cues may be down-weighted or kept separate rather than always fused \cite{guyotoncortical2025}. Implementing such common-cause gating would provide a principled way to prevent inappropriate cross-modal transfer under mismatched inputs. Furthermore, developing principled metrics for dynamically enabling or gating cross-modal connections during learning, i.e., determining when cross-modal learning should occur, represents an important next step. Such mechanisms may improve data-efficiency and robustness while offering a computational perspective on the developmental emergence of cross-modality in biological systems. 

By bridging robotics and neuroscience, this work highlights how bio-inspired computational principles can inform the design of robotic perceptual frameworks capable of robust and efficient behavior in unstructured environments.

\section{Methods}
\label{sec:methods}
\subsection{Experimental setup}
\label{subsec:expsetup}
This study employed the interactive perception setup illustrated in Figure~\ref{fig:experimentalsetuprobot}, featuring a real-world robotic platform designed to validate the proposed approach and benchmark it against baseline methods. The setup comprises two manipulators: a 6-DOF Universal Robots UR5 equipped with a Robotiq two-finger gripper and a 7-DOF Franka Emika Panda arm. Two Contactile tactile sensors~\cite{contactile} were affixed to the inner and outer surfaces of the Robotiq gripper's finger pad, while a Zed2i stereo camera~\cite{zed} was rigidly mounted on the Panda arm. For safety, the maximum operational speed of both arms was limited to 25~mm/s. Ground-truth object pose data was recorded using an OptiTrack motion capture system~\cite{optitrac}. The UR5 arm simulated human-like tactile exploration through two-fingered manipulation, and the Panda arm mimicked neck and eye movements for active visual exploration.

\begin{figure}[!htb]
    \centering
    \includegraphics[width=\textwidth]{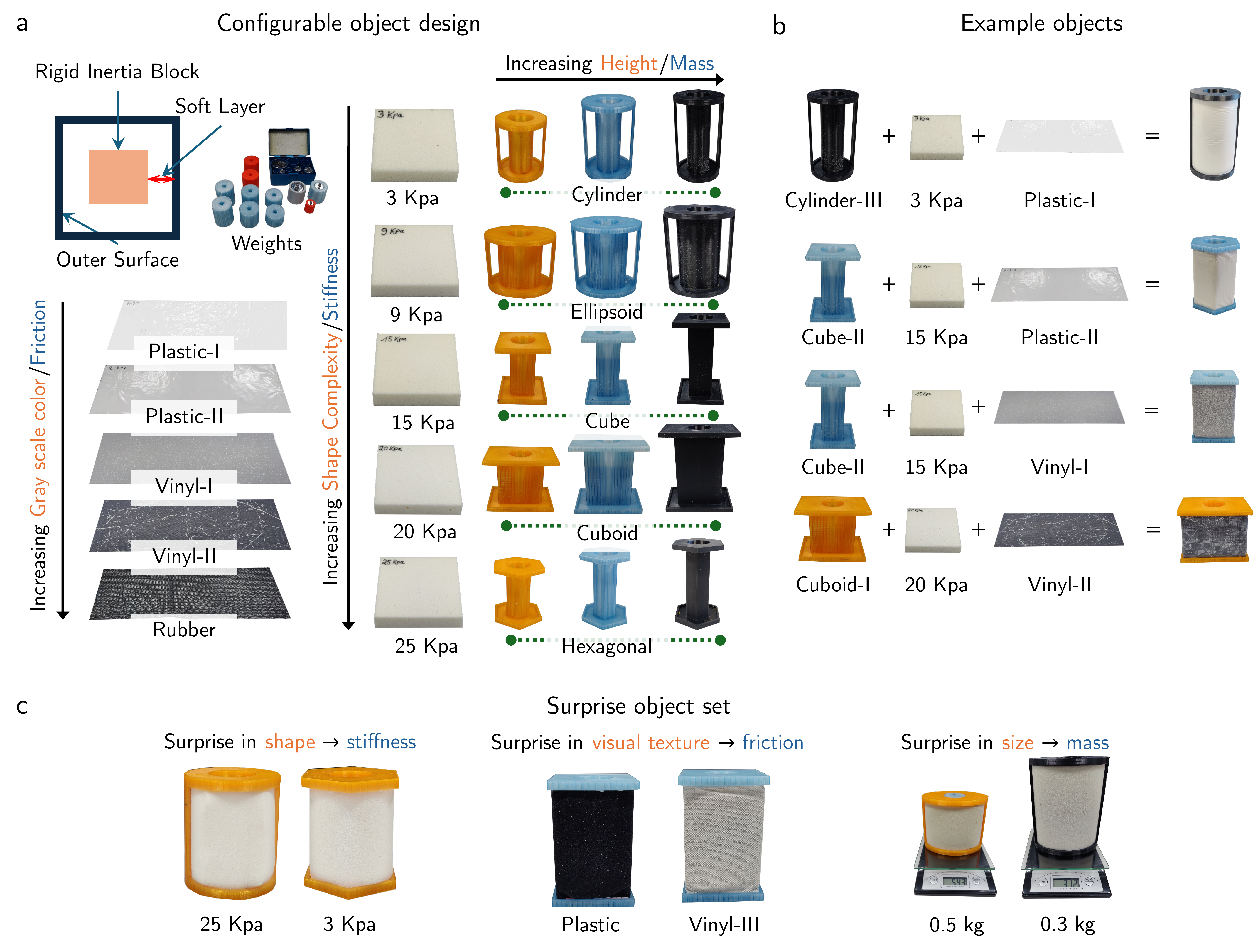}
     \caption{Experimental setup of the configurable non-rigid objects.}
    \label{fig:experimentalsetupcm}
\end{figure}
Everyday or natural objects often exhibit co-varying properties with substantial uncontrolled variability, making it difficult to isolate and thereby hindering controlled evaluation of the cross-modal framework. To enable a systematic evaluation and interpretable assessment, we therefore designed and manufactured configurable non-rigid objects covering wide variation in both intrinsic (mass, stiffness, surface friction) and extrinsic (size, shape, visual texture, color) properties. These objects enable rapid and repeatable control over intrinsic–extrinsic property correlations, far more efficiently than sourcing suitable everyday objects.

Inspired by household items (e.g. food containers, milk cartons), each object comprised: (i) a rigid inner inertial block, 3D printed in one of five prismatic shapes \{\textit{cylinder,\,ellipsoid,\,cube,\,cuboid,\,hexagonal prism}\} and of three height levels \{0.12,\,0.15,\,0.175\}\,m; (ii) a polyurethane foam covering with stiffness values \{3,\,9,\,15,\,20,\,25\}\,kPa; and (iii) an outer textured layer to vary surface friction. Given the difficulty of producing standardized friction surfaces, we employed natural textures and empirically measured the approximate static coefficients measured $\mu \in$ \{0.1,\,0.15,\,0.35,\,0.47,\,0.8\}.

To probe cross-modal priors, we introduced correlations: height with mass, shape with stiffness, and visual texture with surface friction (Figure~\ref{fig:experimentalsetupcm}\,a,b). To decouple shape from mass, inner blocks accommodated modular weights so that all objects of a given height maintained approximately $\pm 100$\,gm same mass regardless of shape. In total $75$ objects were created and $16$ distinct interaction configurations were defined by varying four grip force levels ($F_g$) and four linear/angular velocities ($v_z$/$v_\beta$). Each configuration was repeated three times per object, yielding $48$ trials per object and $3600$ interaction trajectories overall. Additionally, six ``surprise'' objects were created to challenge learned representations: a stiffer cylinder, a softer hexagonal prism, a low friction black object, a high-friction white object, and a short object with greater mass than a taller one (Figure~\ref{fig:experimentalsetupcm}\,c).

\subsection{Initial shape and pose estimation}
\label{subsub:activeshape}
\begin{figure}[!htb]
    \centering
    \includegraphics[width=\textwidth]{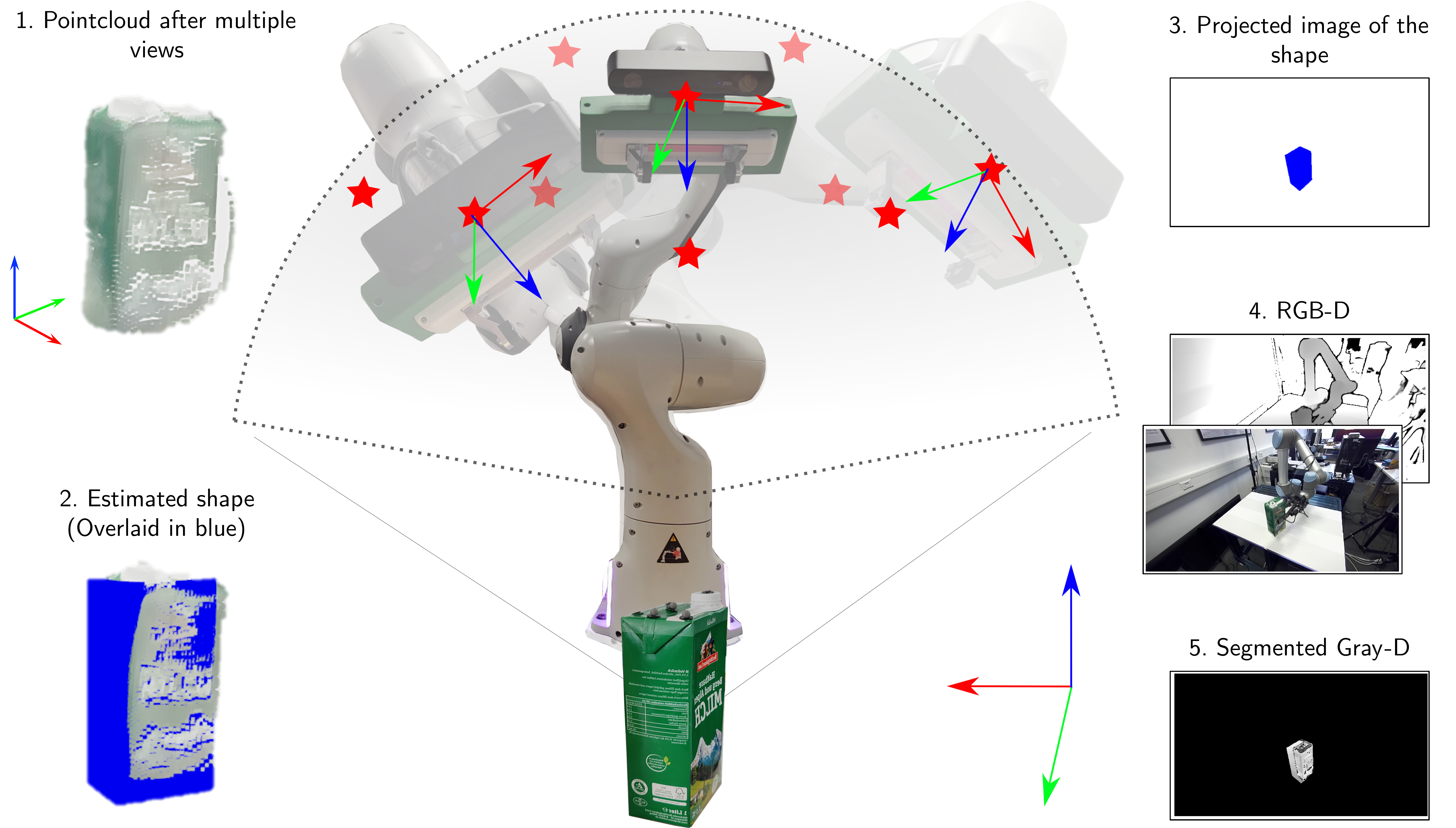}
    \caption{Graphical visualization of the multi-view shape perception and leveraging the estimated shape for segmentation of visual observations.}
    \label{fig:activeshapecm}
\end{figure}
The initial shape and pose estimation was obtained using our previous work on superquadrics~\cite{dutta2024vitract}, which provides a basis for autonomous exploration. Additional preprocessing of raw visual observations was necessary to facilitate visual feature extraction by the encoder, ensure consistent segmentation of the object across successive frames, and avoid overly complex network architectures. Although recent segmentation techniques based on large-scale foundation models could be applied, this was beyond the scope of this work. Instead, we adopted a more efficient strategy: we projected the initially estimated shape onto the image plane using extrinsic camera calibration and updated this mask over time according to the robot’s end-effector pose, assuming stable prehensile contact and negligible object slippage. This approach provided consistent segmentation across frames (Figure~\ref{fig:activeshapecm}).

\subsection{Prehensile interaction}
\label{subsub:prehensile}
We defined a sequence of prehensile interaction primitives to selectively excite object properties: an initial palpation motion to correct for the initial pose $g_{t_0}=[\mathbf{R} \in SO(3), \mathbf{t} \in \mathbb{R}^3]$ from the shape exploration phase,  a grasp with parametrized grip force $F_g$, a lift with linear velocity $v_z$ along the $z$-axis and hand rotation with angular velocity $v_\beta$ about the $y$-axis. Grasping probes stiffness; lifting probes coupled friction-mass-stiffness, and the in-hand rotation help disentangle surface friction from mass.

Formally, the action at time $t$ is defined as $\mathbf{a}_t = (d, v_z, v_\beta)$, where $d$ denotes the commanded distance between the two gripper fingers. As the gripper lacks direct force control, the grip force $gf$ is regulated through a position-based force control scheme. Figure~\ref{fig:experimentalsetuprobot} visualizes the interaction primitive sequence. To determine an appropriate range of grip forces for exploration, a preliminary study was conducted using incipient slip detection, based on tactile signals from the contactile sensor. This provided safe (non-slipping) yet informative interactions, though we note that general slip detection remains a non-trivial problem in robotics \cite{waltersson2024perception}.

\subsection{Data preprocessing and training of CMLF}
\label{subsec:preprocess}
Visual information was collected from a camera with frame rate 30–40\,Hz, and subsequently downsampled to 3\,Hz. Tactile observations were recorded at 600\,Hz, downsampled to 240\,Hz, and restructured into a two-dimensional representation of size $80\times80$ for temporal alignment with the 3\,Hz visual sampling rate while preserving a fine temporal resolution. The raw RGB images of size $640\times240\times3$ were segmented (see Figure~\ref{fig:activeshapecm}), converted to grayscale, and rescaled to a final resolution of $128\times128$. Observations during the initial shape exploration and pre-grasp phases were excluded from analysis. Data collection began with the onset of palpation motion and continued through the grasp–lift–rotation–placement sequence, taking approximately 30\,s per trial. Object placements were manually randomized across trials to yield variability. Ground-truth object pose data was recorded with the OptiTrack motion capture system.

We held out 10\% of the data each for model validation and testing. The proposed approach and baseline techniques were implemented in PyTorch and trained using Adam optimizer with a learning rate of $10^{-5}$. The training process incorporated an annealing strategy to gradually scale the regularization term in the ELBO, promoting stable and effective disentanglement in the learned latent space. Careful hyper-parameter optimization was performed for the learning rate, latent variable dimensionality, and the temperature parameter of the annealing term. 

\begin{figure}[!b]
    \centering
    \includegraphics[width=\textwidth]{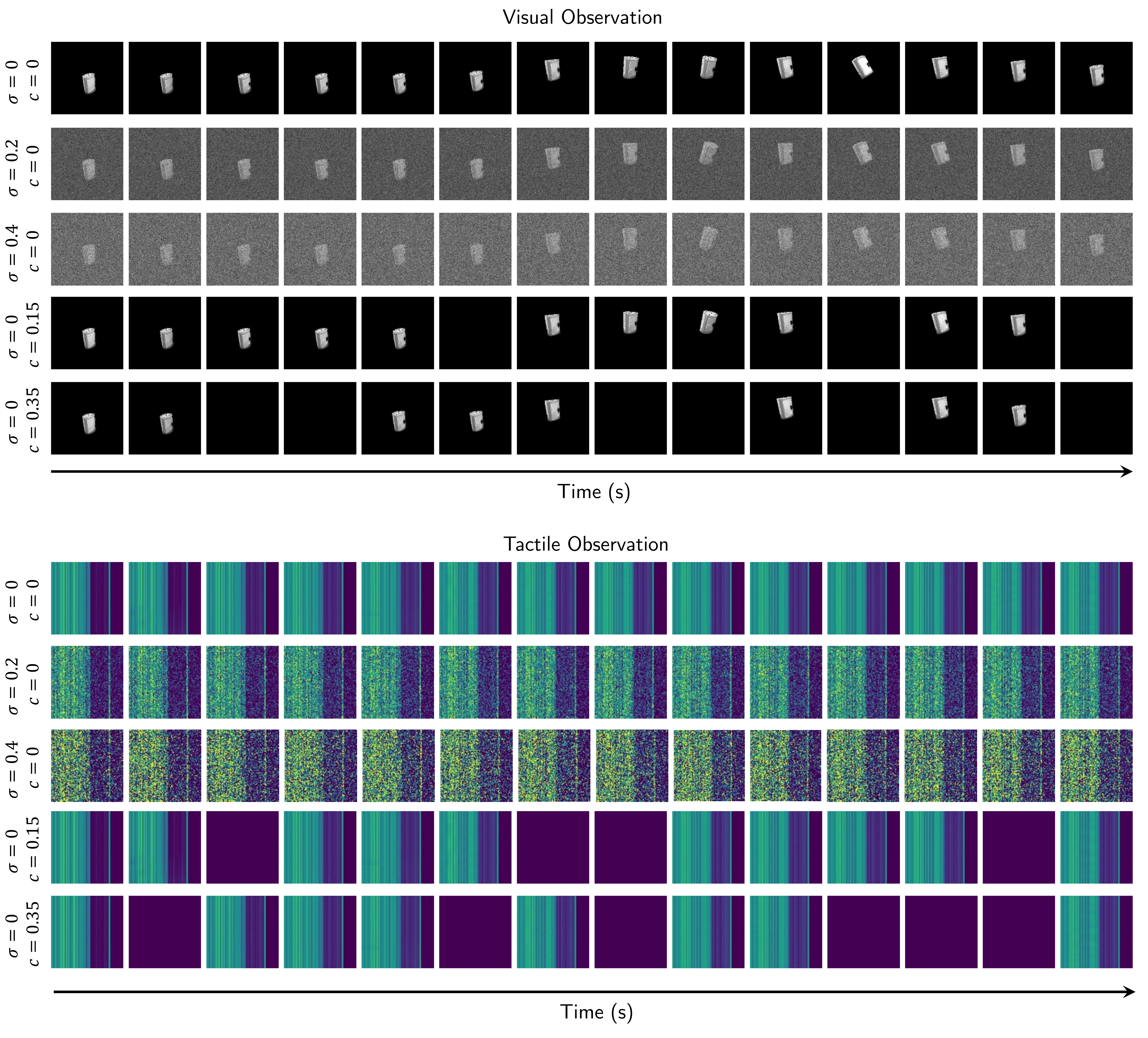}
    \caption{Visualization of the different noise and corruption levels used to evaluate the proposed approach. $\sigma=0.2, 0.4$ denotes the standard deviation of zero mean Gaussian white noise. $c=0.15, 0.35$ denotes the fraction of observations along the trajectory that are replaced with zero values or missing data.}
    \label{fig:cmnoise}
\end{figure}

\subsection{Classification, regression techniques, and statistical analysis}
\label{subsec:classreg}
We evaluated classification performance using logistic regression on learned features ($\mathbf{y}^{V/T}_t$, dimension $n_y=16$) and used UMAP to visualize high-dimensional latent representations. However, direct projection of all 75 object instances into 3D remained challenging, and Euclidean/cosine distances did not yield meaningful insights (Figure~2,  see Supplementary, Section~2). Thus, we aligned the latent space with ground-truth properties following \cite{latentmatters}, allowing analysis of the temporal evolution of physically grounded latent dimensions.

Kernel Ridge Regression (KRR) with a radial basis function (RBF) kernel was used to align the latent representations corresponding to physical parameters (e.g. object properties and pose) using three separate KRR models:
\begin{align}
    f^{V}_y &: \mathbb{R}^{n^{V}_y} \to \mathbb{R}^{3} \quad \text{(visual y-latent $\to$ extrinsic physical properties)} \\
    f^{V}_z &: \mathbb{R}^{n^{V}_z} \to \mathbb{R}^{6} \quad \text{(visual z-latent $\to$ pose)} \\
    f^{T}_y &: \mathbb{R}^{n^{T}_y} \to \mathbb{R}^{3} \quad \text{(tactile y-latent $\to$ intrinsic physical properties)} 
\end{align}
For the baseline approach, a similar set of three functions was trained, with inputs to $f^{V}_y$ and $f^{T}_y$ derived from the shared latent space rather than the modality-specific spaces. Regressors $f^{V}_y$ and $f^{T}_y$ were trained in the final time steps (28–30\,s) of each interaction sequence, then used to predict the aligned ground-truth parameters from samples drawn along the evolving latent trajectories. In contrast, $f^{V}_z$ was trained using a portion of the validation set over the full interaction duration (0–30\,s) to predict time-varying pose. Although the shape was estimated via superquadric fitting, for the present evaluation, we adopted a simplified discrete representation (0-4) corresponding to geometric primitives from \textit{cylinder} to \textit{hexagonal prism}. This abstraction tests whether the self-supervised approach captures relevant shape information despite reduced representation granularity.

Statistical significance was assessed using paired hypothesis tests, reflecting the repeated-measures nature of the experimental design.
Unless stated otherwise, an initial significance level of $\alpha = 0.05$ was used.
To control the family-wise error rate arising from multiple comparisons, we applied Holm–Bonferroni correction within each family of tests (e.g., method-wise comparisons for a given property).
Corrected $p$-values are reported in all figures, with significance indicated as \textbf{*} for $p < 0.05$, \textbf{**} for $p < 0.01$, and \textbf{***} for $p < 0.001$.

\subsection{Perturbation on visuo-tactile observation}
\label{subsec:perturbation}

Figure~\ref{fig:cmnoise} illustrates visual and tactile trajectories under varying levels of input degradation.

\section{Data availability}
A representative subset of the data is available at: \url{https://drive.google.com/drive/folders/1_9xQpeeUF4r76zLRBYKnzWGIbY9OwNoL?usp=sharing}. The complete data set generated and analyzed during this study is not publicly available due to its large size ($\sim$5 TB), but can be obtained from the authors upon reasonable request.

\section{Code availability}
The code supporting this study is publicly available on the GitHub repository: , which can be accessed at: \url{https://github.com/anirvan95/Cross-Modal_LatentFilter.git}

\section{Acknowledgments}
The authors thank Morteza Teymoori for his contribution in the CAD design of the experimental objects. The work described in this paper was carried out at BMW Group AG. It was supported in part by BMW Group and in part by the European Union’s Horizon 2020 research and innovation programme under Grant Agreement No. 861166 (INTUITIVE) and 860949 (SmartNets), and by the European Union’s Horizon Europe programme under Grant Agreement No. 101092096 (PHASTRAC).

\section{Author contributions}
The contributions of the authors are summarized as follows:

\begin{itemize}
\item \textbf{Conceptualization:} A. Dutta, S. Tasciotti, E. Burdet, and M. Kaboli conceived the cross-modal perception problem and the overall research direction.

\item \textbf{Methodology:} A. Dutta led the development and implementation of the Cross-Modal Latent Filter (CMLF), with contributions from C. Cusseddu, A. Li, P. van der Smagt, and M. Kaboli.

\item \textbf{Experiments:} A. Dutta, S. Tasciotti, A. Li, E. Burdet, P. van der Smagt, and M. Kaboli designed the experiments. A. Li and A. Dutta conducted the real-robot experiments.

\item \textbf{Analysis and Writing:} All authors except A. Li contributed to the interpretation of the results and manuscript preparation.
\end{itemize}
All authors reviewed and approved the final version of the manuscript.

\section{Competing interests}
All authors declare no financial conflicts of interest.
{\hypersetup{hidelinks}
\listoffigures{}
\listoftables{}
}

\bibliography{sn-bibliography}

\newpage
\setcounter{section}{0}
\setcounter{figure}{0}
\setcounter{table}{0}

\renewcommand{\thesection}{S\arabic{section}}
\renewcommand{\thefigure}{S\arabic{figure}}
\renewcommand{\thetable}{S\arabic{table}}

\begin{center}
  {\LARGE Supplementary Materials} 
\end{center}

This document accompanies the main manuscript and comprises a supporting derivation and additional results. 
\section{Derivation of Latent Filter}
\label{subsec:appendderiv}
The objective is to model the joint probability $p(\mathbf{o}_{1:T}, \mathbf{s}_{1:T}|\mathbf{a}_{1:T})$, and perform variational inference by maximizing the likelihood of observations. 

\begin{align}
    p(\mathbf{o}_{1:T}|\mathbf{a}_{1:T}) &= \int \!  p(\mathbf{o}_{1:T}, \mathbf{s}_{1:T}|\mathbf{a}_{1:T}) \, d\mathbf{s}_{1:T}
\end{align}
This is intractable! To address this, a variational distribution $q(\mathbf{s}_{1:T})$ is introduced to approximate the true posterior. Under the variational inference scheme, optimized by maximizing the Evidence Lower Bound (ELBO), derived as follows. Starting from the marginal likelihood:
\begin{equation}
    \log p(\mathbf{o}_{1:T} \mid \mathbf{a}_{1:T}) = \, \log \left( \int \! p(\mathbf{o}_{1:T}, \mathbf{s}_{1:T} \mid \mathbf{a}_{1:T}) \, d\mathbf{s}_{1:T} \right) \! ,
\end{equation}
introducing a family of distribution $q(\mathbf{s}_{1:T})$ (using the identity $\int q(\mathbf{s}_{1:T}) \, d\mathbf{s}_{1:T} = 1$), we have:
\begin{equation}
    \log p(\mathbf{o}_{1:T} \mid \mathbf{a}_{1:T}) = \log \mathbb{E}_{q(\mathbf{s}_{1:T})} \left[ \frac{p(\mathbf{o}_{1:T}, \mathbf{s}_{1:T} \mid \mathbf{a}_{1:T})}{q(\mathbf{s}_{1:T})} \right].
\end{equation}
Applying Jensen's inequality gives the variational lower bound:
\begin{equation}
    \log p(\mathbf{o}_{1:T} \mid \mathbf{a}_{1:T}) \geq \mathbb{E}_{q(\mathbf{s}_{1:T})} \left[ \log p(\mathbf{o}_{1:T}, \mathbf{s}_{1:T} \mid \mathbf{a}_{1:T}) - \log q(\mathbf{s}_{1:T}) \right].
\end{equation}
Defining the Evidence Lower Bound (ELBO):
\begin{equation}
    \mathcal{F}_{\textsc{ELBO}}(q) = \mathbb{E}_{q(\mathbf{s}_{1:T})} \left[ \log p(\mathbf{o}_{1:T}, \mathbf{s}_{1:T} \mid \mathbf{a}_{1:T}) \right] - \mathbb{E}_{q(\mathbf{s}_{1:T})} \left[ \log q(\mathbf{s}_{1:T}) \right]
\end{equation}
Maximizing $\mathcal{F}_{\textsc{ELBO}}(q)$ with respect to $q$ thus provides a principled way to approximate the posterior. For dynamic scenarios, a generative model is assumed with an underlying latent dynamical system with: 
\begin{align}
    \label{eq:generativemodelgen}
    p(\mathbf{o}_{1:T}|\mathbf{a}_{1:T}) &=  \int \!\! p(\mathbf{o}_{1:T}|\mathbf{s}_{1:T}, \mathbf{a}_{1:T}) \,p(\mathbf{s}_{1:T}|\mathbf{a}_{1:T}) \, d\mathbf{s}_{1:T} \nonumber \\
    = \int \prod_{t=1}^T \, p(\mathbf{o}_t|\mathbf{o}_{1:t-1}, \mathbf{s}_{t}, &\mathbf{s}_{1:t-1}, \mathbf{a}_{1:t})p(\mathbf{s}_t|\mathbf{o}_{1:t-1}, \mathbf{s}_{t-1}, \mathbf{s}_{1:t-2}, \mathbf{a}_{t}, \mathbf{a}_{1:t-1}) \, d\mathbf{s}_{1:t}
\end{align}
Eq. \ref{eq:generativemodelgen} becomes computationally very expensive with increasing time steps, as with each time step, the conditional variables increase. Therefore, Markov's assumption is used to simplify:
\begin{align}
    p(\mathbf{o}_{t}|\cancel{\mathbf{o}_{1:t-1}}, \mathbf{s}_{t}, \cancel{\mathbf{s}_{1:t-1}}, \cancel{\mathbf{a}_{1:t}}) &= p(\mathbf{o}_{t}|\mathbf{s}_{t}) \nonumber \\
    p(\mathbf{s}_{t}|\cancel{\mathbf{o}_{1:t-1}}, \mathbf{s}_{t-1}, \cancel{\mathbf{s}_{1:t-2}}, \mathbf{a}_{t}, \cancel{\mathbf{a}_{1:t-1}}) &= p(\mathbf{s}_{t}|\mathbf{s}_{t-1}, \mathbf{a}_{t}) \nonumber 
\end{align}
resulting in the following simplified generative model: 
\begin{align}
    p(\mathbf{o}_{1:T}|\mathbf{a}_{1:T}) = \, p(\mathbf{o}_1|\mathbf{s}_1) \!\int \! \prod_{t=2}^T p(\mathbf{o}_t|\mathbf{s}_t) \, p(\mathbf{s}_t|\mathbf{s}_{t-1}, \mathbf{a}_t) \, d\mathbf{s}_t
    \label{eq:generativemodelmarkov}
\end{align}
A further assumption is introduced in the observation model based on the nature of the interaction and the principles of observability of the control theory. This facilitates the analytical computation of the posterior while avoiding the computational expense associated with linearizing the observation model. Accordingly, the latent variable is restructured:
\begin{align}
    p(\mathbf{s}_t) = p(\mathbf{z}_t, \mathbf{y}_t) &= p(\mathbf{z}_t|\mathbf{y}_t) \, p(\mathbf{y}_t)
\end{align}
with directly observable variable $\mathbf{z}_t$ and indirectly observable part as $\mathbf{y}_t$ resulting in the following: 
\begin{align}
   p(\mathbf{o}_{1:T}|\mathbf{a}_{1:T}) = \iint  \prod_{t=1}^T p(\mathbf{o}_t|\mathbf{z}_{t}) \, p(\mathbf{z}_t|\mathbf{z}_{t-1},\mathbf{y}_t, \mathbf{a}_t) \, p(\mathbf{y}_t|\mathbf{y}_{t-1}, \mathbf{a}_{t-1}) \, d\mathbf{y}_{t-1}  d\mathbf{z}_{t-1}
    \label{eq:generativemodelprop}
\end{align}
\begin{figure}[!hb]
    \centering
    \includegraphics[width=0.6\textwidth]{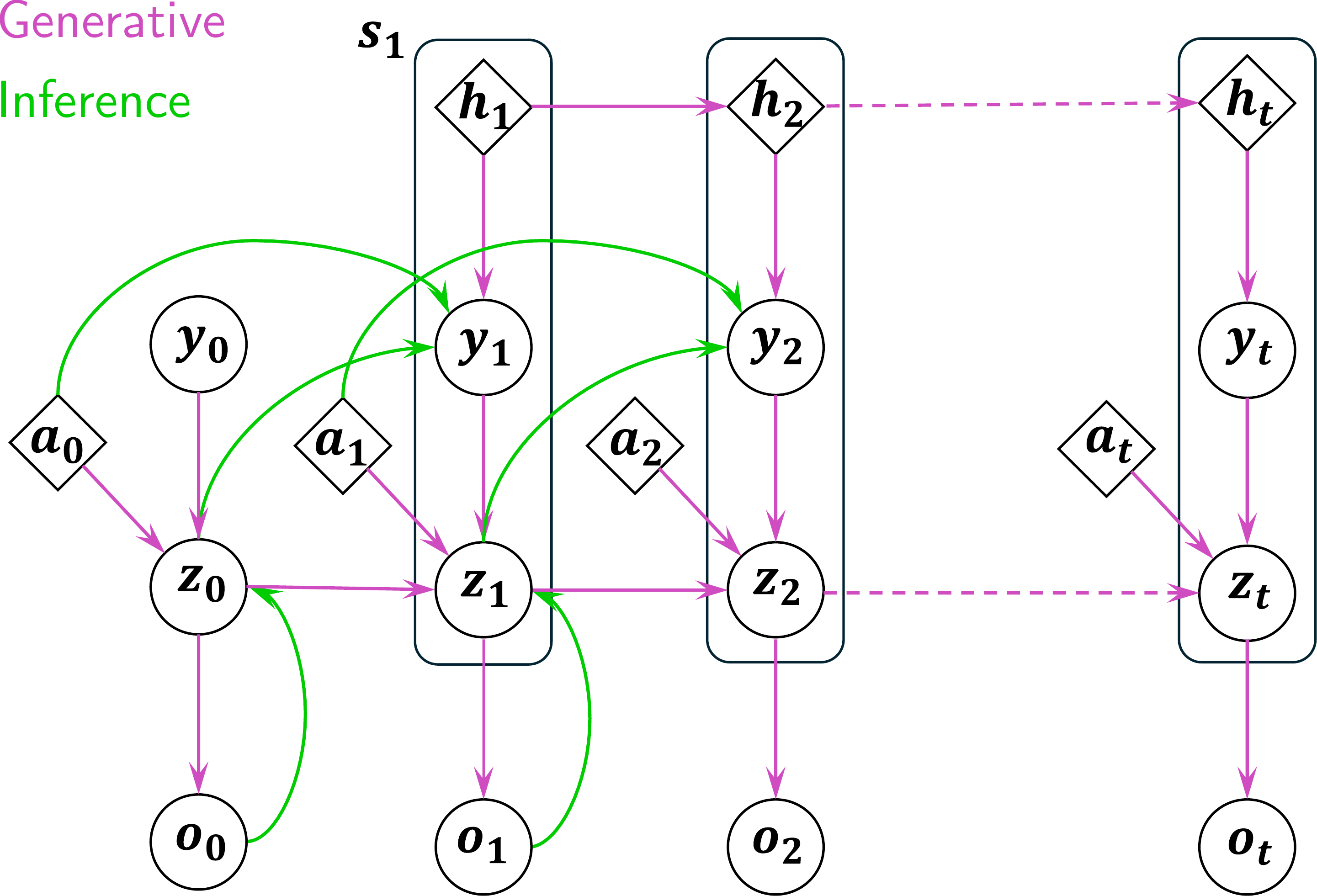}
    \caption{Probabilistic Graphical Model of the Latent Filter}
    \label{fig:placeholder}
\end{figure}
To compute the observation likelihood, variational distribution $q_\theta(\mathbf{z}_{1:T}, \mathbf{y}_{1:T})$ $\sim p(\mathbf{z}_{1:T}, \mathbf{y}_{1:T}|\mathbf{o}_{1:T}, \mathbf{a}_{1:T})$ is introduced. The Evidence Lower Bound Objective (ELBO) for the generative model in Eq. \ref{eq:generativemodelprop} is formulated from \textsc{KL} Divergence inequality:
\begin{align}
    D_{\textsc{KL}} = - \!\!\iint \!q_\theta(\mathbf{z}_{1:T}, \mathbf{y}_{1:T}|\mathbf{o}_{1:T}, \mathbf{a}_{1:T}) 
    \log  \! \left[\frac{p(\mathbf{z}_{1:T}, \mathbf{y}_{1:T}|\mathbf{o}_{1:T}, \mathbf{a}_{1:T}}{q_\theta(\mathbf{z}_{1:T}, \mathbf{y}_{1:T}|\mathbf{o}_{1:T}, \mathbf{a}_{1:T}}\right] d\mathbf{z}_{1:T} \, d\mathbf{y}_{1:T} \geq 0
\end{align}
which results in the following objective function:
\begin{equation}
\log p(\mathbf{o}_{1:T}|\mathbf{a}_{1:T}) \geq \mathbb{E}_{q_\theta}\!\left[ \log p(\mathbf{o}_{1:T}|\mathbf{z}_{1:T}, \mathbf{y}_{1:T}, \mathbf{a}_{1:T}) \right] - \mathbb{E}_{q_\theta} \!\! \left[\log \frac{q_\theta(\mathbf{z}_{1:T}, \mathbf{y}_{1:T}|\mathbf{o}_{1:T}, \mathbf{a}_{1:T})}{p(\mathbf{z}_{1:T}, \mathbf{y}_{1:T}|\mathbf{a}_{1:T})} \right]
\end{equation}
re-introducing the Markov assumption in the regularization term of the ELBO
\begin{align}
    q_\theta(\mathbf{z}_{1:T}, \mathbf{y}_{1:T}  |\mathbf{o}_{1:T}, \mathbf{a}_{1:T}) &= 
    q_{\theta}(\mathbf{z}_1|\mathbf{y}_1) \, q_{\phi}(\mathbf{y}_1) \prod_{t=2}^{T}q_\theta(\mathbf{z}_t, \mathbf{y}_t|\mathbf{z}_{t-1}, \mathbf{y}_{t-1}, \mathbf{o}_{1:t}, \mathbf{a}_{1:t}) \nonumber \\
    q_\theta(\mathbf{z}_t, \mathbf{y}_t|\mathbf{z}_{t-1}, \mathbf{y}_{t-1}, \mathbf{o}_{1:t}, \mathbf{a}_{1:t}) &= \frac{p(\mathbf{o}_t|\mathbf{z}_t) \, q_\theta(\mathbf{z}_t|\mathbf{z}_{t-1}, \mathbf{y}_t, \mathbf{a}_t) \, q_\phi(\mathbf{y}_t|\mathbf{z}_{t-1}, \mathbf{y}_{t-1}, \mathbf{a}_{t-1})}{\iint p(\mathbf{o}_t|\mathbf{z}_t) \, p(\mathbf{z}_t|\mathbf{z}_{t-1}, \mathbf{y}_t, \mathbf{a}_t) \, d\mathbf{z}_{t-1}d\mathbf{y}_t} \nonumber \\
    \sim  \eta &\underbrace{q_{\psi}(\mathbf{z}_t|\mathbf{o}_t) \, q_\theta(\mathbf{z}_t|\mathbf{z}_{t-1}, \mathbf{y}_t, \mathbf{a}_t)}_{q^{filt}(\mathbf{z}_t)} \, q_\phi(\mathbf{y}_t|\mathbf{z}_{t-1}, \mathbf{y}_{t-1}, \mathbf{a}_{t-1})
\end{align}
where the denominator $\eta$ is the normalization factor and resulting in the simplified ELBO as:
\begin{align}
    \mathcal{F}_{\textsc{ELBO}}(\theta, \phi)  = \,\mathbb{E}_{q_\phi(.)} \!\! \left[ \sum_{t=1}^{T} \log p(\mathbf{o}_t|\mathbf{z}_t)\right] 
    - &\sum_{t=2}^{T}{\textsc{KL}}[ q^{filt}(\mathbf{z}_t) || p(\mathbf{z}_{t}|\mathbf{z}_{t-1}, \mathbf{y}_{t}, \mathbf{a}_{t}) \nonumber \\
    -  \sum_{t=2}^{T}{\textsc{KL}}[ &q_\phi(\mathbf{y}_t|.)||p(\mathbf{y}_t| \mathbf{a}_{t})]
\end{align}
The extension of the above is the proposed cross-modal latent filter, with separate visual and tactile spaces and additional prior connection integrated. The updated ELBO is provided in the main text. 
\newpage
\section{Additional results}
\label{subsec:appendres1}
\begin{figure}[!htb]
    \centering
    \includegraphics[width=\textwidth]{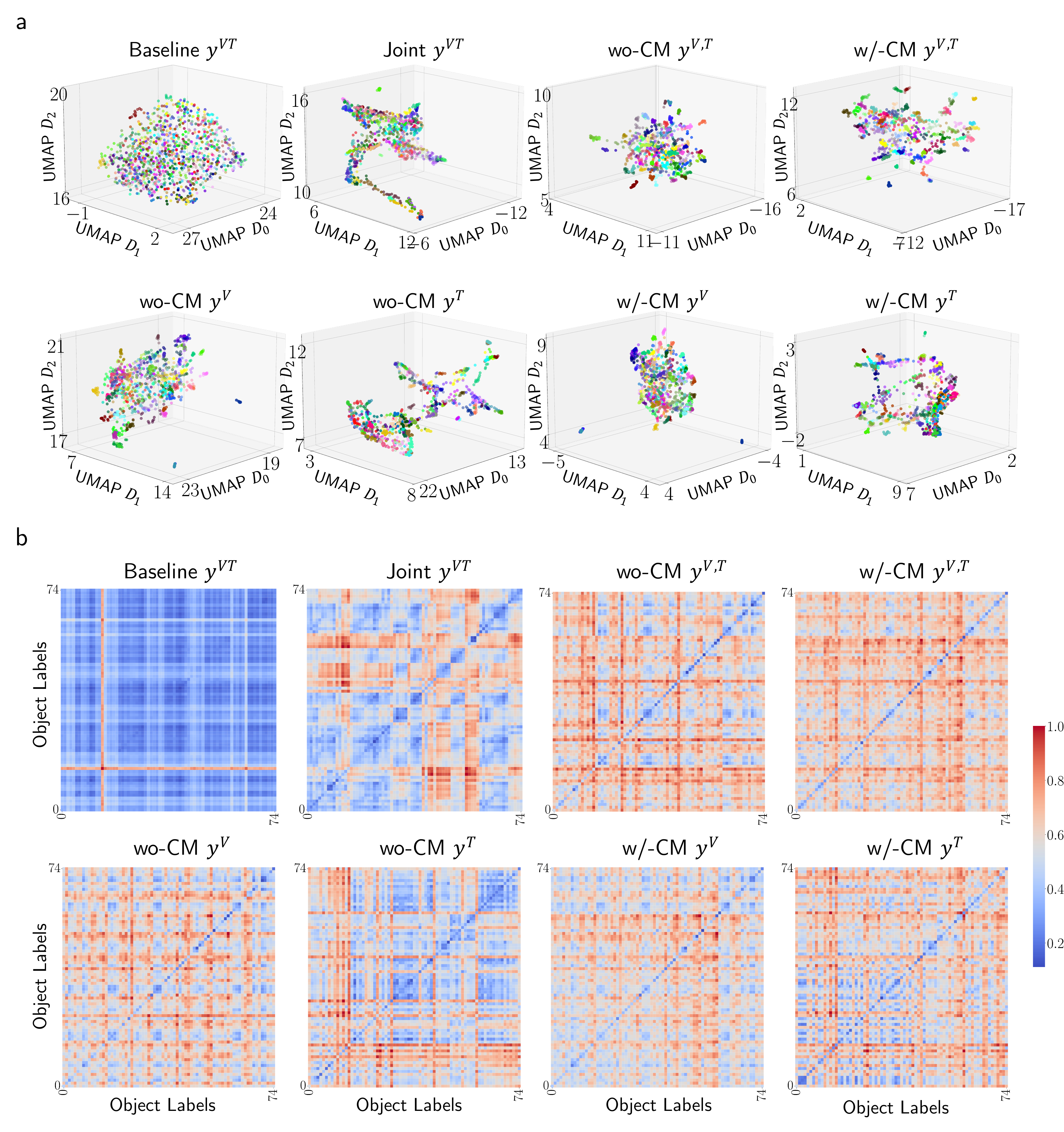}
    \caption{Qualitative analysis of latent space}
    \label{fig:latentumadistance}
\end{figure}
\begin{figure}[!htb]
    \centering
    \includegraphics[width=\textwidth]{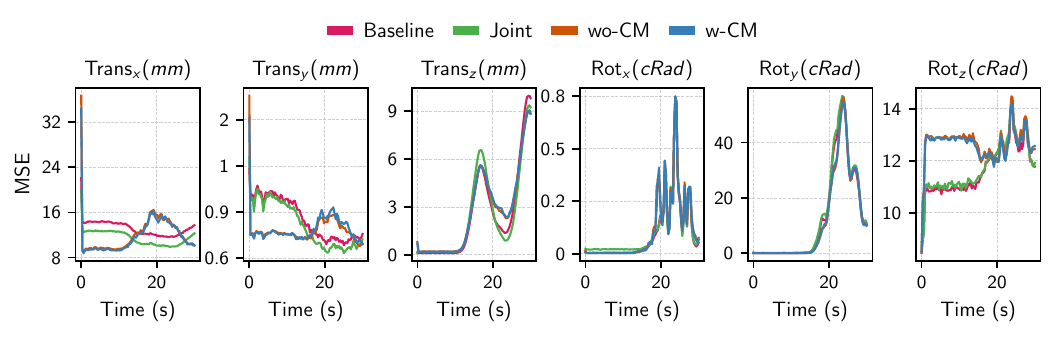}
    \caption{Temporal evolution of the NMSE error of pose estimation, with the shaded region depicting $\pm 0.1$ standard deviation, representing the variability of the mean error. }
    \label{fig:poseregression}
\end{figure}
\end{document}